\RequirePackage{fix-cm}
\documentclass[11pt,letterpaper]{article}
\usepackage[margin=1in]{geometry}
\usepackage{iftex}
\ifPDFTeX
  \usepackage[T1]{fontenc}
\else
  \usepackage{fontspec}
\fi
\usepackage{amsmath,amssymb,amsthm,mathtools}
\usepackage{booktabs,array,enumitem}
\usepackage[round,authoryear]{natbib}
\usepackage{setspace,microtype}
\usepackage[colorlinks=true,linkcolor=blue,citecolor=blue,urlcolor=blue]{hyperref}
\hypersetup{pdftitle={The Cost of Working Memory in Streaming Mean Estimation},
  pdfauthor={Zeyu Wu, Zihao Wang, Minghua Deng}}
\newtheorem{theorem}{Theorem}[section]
\newtheorem{proposition}[theorem]{Proposition}
\newtheorem{lemma}[theorem]{Lemma}
\newtheorem{corollary}[theorem]{Corollary}
\theoremstyle{definition}

\newcommand{\E}{\mathbb E}
\newcommand{\Pp}{\mathbb P}
\newcommand{\R}{\mathbb R}
\newcommand{\N}{\mathbb N}
\newcommand{\Rstr}{\mathcal R^{\mathrm{str}}}
\newcommand{\Rterm}{\mathcal R^{\mathrm{term}}}
\newcommand{\Rfree}{\mathcal R^{\mathrm{free}}}

\DeclareMathOperator{\tr}{tr}

\DeclareMathOperator{\Bern}{Bern}
\newcommand{\ind}{\mathbf 1}
\numberwithin{equation}{section}
\setlist[itemize]{topsep=4pt,itemsep=2pt}
\setlist[enumerate]{topsep=4pt,itemsep=2pt}
\title{The Cost of Working Memory in Streaming Mean Estimation}
\author{Zeyu Wu\quad Zihao Wang\quad Minghua Deng\thanks{Corresponding author.}\\[0.4em]
  \normalsize School of Mathematical Sciences, Peking University}
\date{}
\begin{document}
\setstretch{1.06}
\maketitle
\begin{abstract}
A streaming estimator must compute from the evidence that fits in memory.
We determine when storing an accurate mean is easier than computing it.
For Gaussian mean estimation over the Euclidean unit ball in $\R^d$, we
establish the minimax rate under squared error with $K$ persistent states, a known
horizon, and a public time index:
\[
\min\left\{1,\ K^{-2/d}+\frac dn
\left(1+\frac{\log(ed)}{1+(\log K)/d}\right)\right\}.
\]
Unrestricted computation followed by $K$-level encoding has rate
$\min\{1,d/n+K^{-2/d}\}$. At a fixed sufficiently small error, statistically
optimal sample size therefore requires order $d\log d$ working bits, while
order $d$ bits suffice to store the answer. The lower bound tracks repeated
state compression through an entropy potential obtained by Gaussian smoothing. A matching
construction combines a multiscale modular counter with workspace reused
across coordinate groups. In one dimension, it attains the terminal rate
$n^{-1}+K^{-2}$. Any fixed confidence requirement that the realized loss
not exceed the zero baseline preserves these rates. Increasing confidence
creates a further memory requirement: at fixed nontrivial risk in every
fixed dimension, failure probability $e^{-\gamma n}$ requires and permits
order $\log n$ working states for sufficiently small fixed $\gamma>0$.
In the scalar problem, a fixed terminal alphabet suffices under the same
protection. These bounds separate the memory needed to represent an answer
from the memory needed to accumulate evidence and validate a decision.
\end{abstract}

\section{Introduction}\label{sec:introduction}

Streaming estimation supports decisions from continuously arriving data, as
in quantitative trading, where online algorithms update statistics of
liquidity and volatility \citep{loveless2013}. A related need to process long
streams with bounded retained information arises in language model inference,
where the attention cache must remain manageable as the sequence grows
\citep{xiao2024}.
A streaming estimator must extract information before each observation is
discarded. Its retained state is the evidence available for the next update,
so a memory constraint can change statistical accuracy as well as storage
cost. A running average appears to need just one vector, but this description
counts real numbers rather than bits. With a finite bit budget, the same
state must represent the current estimate and retain enough evidence to
improve it. The space needed for an accurate final answer may therefore
understate the space needed to compute that answer.

We ask when answer storage suffices for statistically efficient computation.
We study independent observations $X_t\sim N_d(\mu,I_d)$ with
$\|\mu\|\le1$, a known sample size $n$, and at most $K=2^b$ persistent
states at each time. The index $t$ is public, allowing scheduled measurements
without a stored clock. All information obtained from the data passes through the
current state. Comparing this model with unrestricted computation followed
by a $K$-level encoding isolates the statistical cost of working memory.

The main result separates that cost from sampling and answer representation:
\begin{equation}\label{eq:intro-rate}
\underbrace{K^{-2/d}}_{\text{answer representation}}
+\underbrace{\frac dn}_{\text{sampling}}
+\underbrace{\frac{d^2\log(ed)}{n(d+\log K)}}_{\text{working memory}},
\end{equation}
with the total truncated at a constant. The last term is necessary for
randomized procedures with arbitrary joint states and transitions that
vary with time. A deterministic construction attains the same rate. Thus the
memory budget determines how efficiently the stream can be converted into
an estimate, even when the final answer already fits in that budget.

At a fixed sufficiently small target risk, the difference is an order
$\log d$ factor in bits: order $d$ bits represent an adequate answer,
whereas achieving that risk with order $d$ samples requires order
$d\log d$ working bits. Corollary~\ref{cor:main-memory} gives the precise
relationship between sample size and memory. The separation also persists when accuracy
improves with dimension: Corollary~\ref{cor:main-separation} gives sequences
for which both risks vanish but their ratio diverges. Working memory is
therefore a statistical resource beyond the precision of the output.

The matching bounds explain how this resource is used. In one dimension,
a modular counter measures the mean at decreasing frequencies. Its counter
shrinks as the estimate label grows, keeping their joint state space within
budget and attaining $n^{-1}+K^{-2}$. In higher dimensions, completed
coordinate estimates are encoded jointly while the workspace is reused
for the next group. The lower bound tracks the information lost at each
state update through an entropy potential obtained by Gaussian smoothing. Its cumulative
accounting yields the last term of \eqref{eq:intro-rate} for arbitrary
joint states, matching the workspace allocation in the construction.

Confidence introduces a second reason to retain working memory. We require
the terminal estimate to have loss greater than that of zero with probability
at most $\delta$, uniformly over the mean. Fixed $\delta$ changes the rates
only by constants. As $\delta$ decreases, validating a candidate introduces
an asymmetric testing problem with its own tradeoff between sample size and the number of states. At fixed
nontrivial risk, failure probability $e^{-\gamma n}$ requires and permits
order $\log n$ states, or $\log_2\log n+O(1)$ bits, for sufficiently small
fixed $\gamma>0$. We establish this law first in one dimension and then
extend it to every fixed dimension by reusing the validation workspace
across coordinates. Sections~\ref{sec:dimension}--\ref{sec:methods}
develop the estimation cost; Section~\ref{sec:confidence} explains the
additional cost of confidence.

\subsection{Related work}

Inference with finite memory depends on both the transition model and the sample
size. Small machines whose transitions vary with time can accumulate arbitrarily reliable
testing evidence \citep{cover1969}, and mean estimators with transitions that vary over time achieve
prescribed asymptotic accuracy for bounded observations \citep{wagner1972}.
For Gaussian observations, machines with stationary transitions can approach the risk of
an optimal quantizer with finitely many levels as the sample size tends to infinity
\citep{hellman1974}. Optimality at finite sample sizes was also studied for binary
Bayes testing between two known distributions \citep{cover1976}.
Our scalar result gives a deterministic construction that attains
$\min\{1,n^{-1}+K^{-2}\}$ jointly in sample size and state budget, under a
public clock and fixed horizon. Deterministic Bernoulli estimation with stationary transitions has
asymptotic risk of order $K^{-1}$ \citep{berg2021}; these modeling
distinctions are surveyed in \citet{berg2024}.

The distinction between the memory needed to represent a solution and
the memory needed to learn it also motivates frameworks connecting
learning with limited memory to statistical queries \citep{steinhardt2016}.
Lower bounds for protocols with information constraints, including online
algorithms with bounded memory, establish separations between sample size
and memory for detecting biased coordinates and estimating sparse covariance \citep{shamir2014}.
We resolve this tradeoff for the Gaussian location family through matching
bounds in $n,d,K$, measured against a benchmark based on terminal encoding.

Quantized estimation measures the accuracy lost when an estimate must have
a short description. Sharp asymptotic tradeoffs are available for Gaussian
sequence models over Euclidean balls \citep{zhu2014}. Our terminal benchmark
permits the same access to all observations before encoding; the additional term in
\eqref{eq:intro-rate} measures the cost of imposing the alphabet bound at
every update.

Estimation under communication constraints imposes a different restriction: each
observation or device sends a limited message, while a central estimator may
retain the transcript. Identities for conditional scores and Fisher information
give powerful lower bounds in this setting
\citep{barnes2020,barnes2021,acharya2022}. The sign reduction from a Gaussian
location observation to a Bernoulli observation is also used in that
literature \citep{acharya2022}. Our lower bound must additionally account for
recompression of the old state after every new observation. It does so by
telescoping over time an entropy potential obtained by Gaussian smoothing. Gaussian
smoothing and its properties in information theory have a broader history
\citep{guo2005,polyanskiy2016}; the role of smoothing here is to charge the
loss caused by repeated compression by conditional expectations.

Streaming estimation in high dimensions is itself well established. Robust
mean estimation admits computationally efficient algorithms that use a single pass with
space nearly linear in dimension, up to logarithmic factors
\citep{diakonikolas2022}. Our clean Gaussian experiment instead resolves the
risk as a function of an arbitrary total state budget, including the
logarithmic cost of working memory at statistically optimal sample size.
Sample lower bounds under finite memory have also been developed for noiseless
Gaussian regression \citep{openai2026}. That experiment supplies exact
linear equations for an unknown direction. The noisy location experiment
studied here instead requires averaging independent perturbations, and its
cost of working memory is measured against a terminal quantization benchmark.
The upper bound uses a scalar procedure with finitely many states and
encodes its outputs jointly inside a Euclidean ball. Its modular measurement also has a conceptual
relative in modulo quantization with side information
\citep{mayekar2021}; here the side information is an evolving phase label and
both the label and counter are charged to the same state budget.

Confidence guarantees for realized loss improvement provide comparisons
against an incumbent estimator \citep{trippe2024}. Taking zero as the
incumbent, we determine how baseline protection changes the sample and
working memory requirements at fixed target risk.

\section{Statistical experiment and memory model}\label{sec:model}

Let $B_2^d=\{\mu\in\R^d:\|\mu\|\le1\}$, where $\|\cdot\|$ is Euclidean norm.
We observe $X_1,\ldots,X_n$ independently from $N_d(\mu,I_d)$ and evaluate
an action $w\in\R^d$ by
\[
R_\mu(w)=\tfrac12\|w-\mu\|^2.
\]
The zero action has worst-case loss $1/2$.

\paragraph{Streaming procedures.}
A procedure starts in a state $S_0$ whose law is independent of the parameter and updates
\begin{equation}\label{eq:state-update}
S_t\sim Q_t(\,\cdot\mid S_{t-1},X_t),\qquad S_t\in\{1,\ldots,K\}.
\end{equation}
The final estimate is decoded from $S_n$, possibly using fresh independent
randomness. The kernels may depend on the public quantities $n,d,K,\delta,t$,
but not on $\mu$. They can involve arbitrary measurable computation within a
step. All information retained between observations is part of $S_t$. This
includes any counter, scale, label, or random seed that depends on the data.
Fresh independent randomization is allowed without storing a seed.
A complete vector $X_t$ arrives at each step, and $S_t$ is the sole record
carried to the next observation.

The time index and fixed program are free. Predetermined block and layer
boundaries are therefore recognized without storing a clock. Within each
block, observations are processed individually and discarded. The resource
measured here is persistent memory, with unrestricted computation within
each update, including codebook operations.

\paragraph{Baseline protection.}
A procedure is uniformly $\delta$-safe if
\begin{equation}\label{eq:safety}
\sup_{\mu\in B_2^d}
\Pp_\mu\{R_\mu(\widehat w)>R_\mu(0)\}\le\delta.
\end{equation}
This guarantee controls the realized loss relative to zero, while the
minimax criterion below averages the loss. At $\mu=0$, safety requires
returning zero with probability at least $1-\delta$.

Let $\mathcal A^{\rm str}_{n,K,\delta}(d)$ denote the procedures satisfying
\eqref{eq:state-update} and \eqref{eq:safety}. Define
\begin{equation}\label{eq:minimax}
\Rstr_{n,K,\delta}(d)
=\inf_{\widehat w\in\mathcal A^{\rm str}_{n,K,\delta}(d)}
\sup_{\mu\in B_2^d}\E_\mu R_\mu(\widehat w).
\end{equation}
We write $\Rfree_{n,K}(d)$ for the analogous streaming risk with no safety
constraint. All stated lower bounds for fixed confidence also apply to
$\Rfree_{n,K}(d)$.

\paragraph{Terminal compression.}
For the terminal benchmark, an encoder may use all observations and
unrestricted working memory, but sends one of $K$ labels to the decoder.
The final action must satisfy \eqref{eq:safety}. Its minimax risk is
$\Rterm_{n,K,\delta}(d)$. Both models impose the same restriction on the final
label; only streaming imposes it on every intermediate state.

We take $K=2^b$, with $b\ge0$ an integer. Bounds for other integer alphabet
sizes follow by rounding with resources changing by at most a constant factor. The
notation $a\asymp_\delta b$ means that their ratio is bounded above and below
by positive constants depending only on $\delta$. Unsubscripted logarithms
are natural, and $\log_2$ denotes the logarithm to base two.

\section{Minimax risk and memory requirements in high dimensions}
\label{sec:dimension}

We first give the benchmark that accounts only for representation of the
answer. Its two terms are sampling error and the squared covering
radius of a $K$-point codebook.

\begin{theorem}[Terminal compression]\label{thm:main-terminal}
For every fixed $0<\delta<1$, uniformly in integers $n,d\ge1$ and $b\ge0$,
\[
\Rterm_{n,2^b,\delta}(d)
\asymp_\delta \min\{1,d/n+2^{-2b/d}\}.
\]
The lower bound holds without the safety requirement and allows randomized
encoding and decoding.
\end{theorem}

The streaming problem charges the same state budget at every update.
The resulting risk includes the cost of retaining evidence for future
observations.

\begin{theorem}[Optimal streaming risk]\label{thm:main-highdim}
Put
\begin{equation}\label{eq:Psi}
\Psi(n,d,K)=\min\left\{1,K^{-2/d}+\frac dn
\left[1+\frac{\log(ed)}{1+(\log K)/d}\right]\right\}.
\end{equation}
For every fixed $0<\delta<1$ there is a finite $C_\delta$ such that
\begin{equation}\label{eq:main-highdim}
\frac1{256}\Psi(n,d,K)
\le \Rfree_{n,K}(d)
\le \Rstr_{n,K,\delta}(d)
\le C_\delta\Psi(n,d,K)
\end{equation}
for all integers $n,d\ge1$ and $K=2^b$, $b\ge0$.
The upper bound is achieved by a deterministic procedure whose transitions vary with time.
\end{theorem}

The factor
\[
1+\frac{\log(ed)}{1+(\log K)/d}
\]
quantifies the sample penalty for limited working memory. If the available
bits per coordinate are of order $\log d$, the factor is bounded. If they
remain bounded, it is of order $\log d$. Increasing the number of states can
therefore improve sample efficiency even after the answer has enough
representational precision.

\begin{corollary}[Sample efficiency requires working bits]
\label{cor:main-memory}
Fix $Q>0$ and let $0<r\le256^{-2}$. If a streaming procedure has worst-case
risk at most $r$ using $d\le n\le Qd/r$ observations, then
\begin{equation}\label{eq:memory-necessary}
\log K\ge\max\left\{\frac d4\log\frac1r,
\frac{d\log(ed)}{256Q}-d\right\}.
\end{equation}
Conversely, for each fixed $\delta$, risk at most $r$ is achievable with
$O_\delta(d/r)$ samples and
$O_\delta\bigl(d\max\{\log(1/r),\log(ed)\}\bigr)$ persistent bits.
\end{corollary}

For fixed sufficiently small $r$ and sufficiently large $d$, this gives a
requirement of order $d\log d$ for working memory at statistically optimal sample
size. The terminal benchmark needs only order $d$ bits at that same fixed
risk. For decreasing $r$, both problems must also pay the cost of storing the answer, $d\log(1/r)$; the corollary identifies when that cost dominates the
extra workspace.

\begin{corollary}[A separation with vanishing risks]
\label{cor:main-separation}
Fix $0<\delta<1$. For $d\ge3$, set
\[
n_d=\lceil d(\log d)^2\rceil,\qquad
b_d=\lceil d\log_2\log d\rceil,\qquad K_d=2^{b_d}.
\]
Then, as $d\to\infty$,
\[
\Rterm_{n_d,K_d,\delta}(d)\asymp_\delta\frac1{(\log d)^2},
\qquad
\Rstr_{n_d,K_d,\delta}(d)\asymp_\delta
\frac1{\log d\,\log\log d}.
\]
Their ratio is of order $\log d/\log\log d$.
\end{corollary}

Thus the separation is visible even when both procedures consistently
estimate the mean. A final codebook captures the desired precision, while
repeated compression slows the accumulation of evidence.

Theorems~\ref{thm:main-terminal}--\ref{thm:main-highdim} and both corollaries
are proved in Appendix~\ref{app:highdim}. The next two sections explain the
construction and information bound that produce the matching rate.

\section{Scalar minimax risk and an optimal construction}
\label{sec:scalar}

\begin{theorem}[Scalar equivalence]\label{thm:main-scalar}
For every fixed $0<\delta<1$,
\[
\Rstr_{n,K,\delta}(1)\asymp_\delta
\Rterm_{n,K,\delta}(1)\asymp_\delta
\min\{1,n^{-1}+K^{-2}\}.
\]
A deterministic streaming construction attains the upper bound.
\end{theorem}

In one dimension, the term for working memory in \eqref{eq:Psi} is absorbed by
$n^{-1}$. The constructive content is more informative: an alphabet of
order $M$ states can obtain squared error of order $M^{-2}$ from order
$M^2$ observations. Storing an exact count of those observations would
require order $M^2$ states. The construction avoids that count by retaining
only a modular remainder.

\subsection{Modular measurements and scheduled layers}

For $X_t\sim N(\mu,1)$, define $Y_t=\ind\{X_t\ge0\}$. Then
$Y_t\sim\Bern(p)$ with $p=\Phi(\mu)\in[a,1-a]$, where $a=\Phi(-1)$.
The inverse link is uniformly Lipschitz on this interval. It is therefore
enough to estimate $p$ to squared accuracy $M^{-2}$. Gaussianity enters this
upper bound through the link; the mechanism processes Bernoulli data using finitely many states.

Choose a dyadic resolution $M=2^J$. The algorithm has $J$ layers, determined
before data collection from the resource budgets. At layer $j$, the
frequency is $f_j=M/2^j$.
A measurement block consists of $Af_j^2$ consecutive observations, with a
fixed numerical constant $A$. Throughout the block, the state contains
\begin{equation}\label{eq:mod-counter}
C\leftarrow(C+Y_t)\bmod(Af_j).
\end{equation}
At its end, $C/(Af_j)$ estimates $f_jp$ modulo one with a fixed small error
probability: its fluctuation scale is of order $A^{-1/2}$, independently
of $f_j$.

The old label approximates $2f_jp$ modulo one. If it is $r$, there are two
possible lifts to the current frequency, $r/2$ and $(r+1)/2$, on the unit
circle. The block measurement votes for the closer lift. Decreasing the
frequency resolves a new ambiguity while preserving information from the
previous layer. Each measurement block uses fresh samples and resets the
counter, and the public clock determines the block and layer boundaries.

\subsection{Amplification without storing votes}

Each layer uses several rounds, each containing several independent block
votes. The algorithm retains only two bits $(F,H)$. The bit $F$ is the
current candidate. At the start of a round, $H=1$; a vote agreeing with
$F$ changes $H$ to zero. At the end of the round, $F$ is flipped exactly
when $H$ is still one, that is, when every vote opposed it. The list of
votes is never stored. This device is closely related to the tests with finite memory and transitions that vary with time of \citet{cover1969}.

When each block vote is correct with probability at least $0.99$, using
$\ell_j$ votes per round and $T_j$ rounds reduces the decision error to
$\eta_j=2^{-3j}/16$. The inherited phase determines when this condition
holds. Appendix~\ref{app:scalar} gives the explicit schedule and the
recovery argument for the remaining layers.

\subsection{The state budget and error accounting}

At layer $j$, the old label has $2^{j-1}$ values, the counter has
$Af_j$ values, and the amplifier has four states. Their joint state space
therefore has size
\begin{equation}\label{eq:scalar-state-count-main}
2^{j-1}\cdot Af_j\cdot4=2AM.
\end{equation}
The label grows at exactly the rate at which the counter shrinks.
This exchange of capacity between state components is the allocation mechanism.
The public clock supplies the scheduled round and block indices, so they
require no additional memory.

The decreasing block lengths offset the increasing reliability of later
decisions. In fact, $A\sum_{j=1}^J f_j^2\ell_jT_j\le32AM^2$.
The error analysis charges an erroneous layer according to the final
displacement it can cause. Later accurate
layers recover from earlier errors up to the corresponding spatial scale.
The geometrically decreasing $\eta_j$ make the weighted sum of these losses
of order $M^{-2}$. Appendix~\ref{app:scalar} gives the recovery argument.

Choosing $M$ as the largest admissible power of two under the state and
sample budgets gives $n^{-1}+K^{-2}$. Safety at fixed confidence is then
obtained by shrinking the decoded estimate toward zero. If $q$ has
uniform mean squared error at most $v$, radial soft thresholding at
$\sqrt{v/\delta}$ gives a $\delta$-safe action with risk of order
$(1+\delta^{-1})v$ and requires no new persistent states
(Lemma~\ref{lem:radial-safety}). Thus fixed baseline protection changes
constants but not the rates in one or several dimensions.

\section{Information bounds and workspace allocation}\label{sec:methods}

Running $d$ scalar engines simultaneously stores $d$ working counters.
Running them one after another saves states but discards most coordinates
of each vector observation. The optimal construction balances these costs,
and the lower bound shows that no joint encoding can avoid their statistical effect.

\subsection{Information accumulation under a state constraint}

Let $p_{t,s}(\mu)=\Pp_\mu(S_t=s)$ and
$A_t=\nabla_\mu\log p_{t,S_t}(\mu)$ be the score of the current state.
With $Z_t=X_t-\mu$, the identity for the conditional score is
\begin{equation}\label{eq:score-recursion-main}
A_t=\E_\mu[A_{t-1}+Z_t\mid S_t].
\end{equation}
Every observation adds a fresh Gaussian score, and the next state keeps
only its conditional mean. The score variance
$V_t=\E_\mu\|A_t\|^2$ is the trace of the state Fisher information.
It need not increase: an update may erase information. The proof keeps
these potentially negative increments throughout the entropy accounting.

We smooth the distribution of $A_t$ by an independent Gaussian. Its entropy
above the smoothing noise is bounded by $H(S_t)\le\log K$. An
entropy comparison for conditional expectations bounds the loss from compression by
the associated squared error. Choosing the smoothing variance
according to remaining time and telescoping gives a harmonic information
budget. Freezing the terminal state over a longer horizon and summing by
parts converts this budget into
\begin{equation}\label{eq:fisher-main}
\tr\mathcal J_n(\mu)
\le \frac{2n(2\log K+d)}{\log(n+2)}.
\end{equation}
The numerator in \eqref{eq:fisher-main} measures the information that the
state budget can retain; the logarithmic denominator comes from compression
across time. The van Trees inequality converts
this information bound into risk through integration by parts. For sample sizes at least the dimension,
the logarithm of the horizon supplies the dimension dependence in the
term for working memory. At smaller sample sizes, the sampling lower bound is already of
constant order. Combining these bounds with the codebook lower bound proves
the lower half of Theorem~\ref{thm:main-highdim}. The argument treats the
whole state jointly and imposes no coordinatewise architecture.

\subsection{Reusing workspace across groups}

The upper bound gives stored and active coordinates different roles.
Completed coordinates are rounded and encoded jointly as a point in a
Euclidean ball. A new group uses parallel copies of the scalar engine on
a fresh segment of the stream. Its counters are then discarded and its
rounded answers join the stored label. The same workspace is reused for
the next group. Each coordinate estimate is based on arriving samples;
a group boundary never gives access to old observations.

Joint encoding matters. Encoding every coordinate in a separate interval that covers its full range incurs avoidable factors of $\sqrt d$. With grid spacing
$\varepsilon/\sqrt d$, a bounded ball in $k$ completed coordinates contains
at most $(C\varepsilon^{-1}\sqrt{d/k})^k$ grid points. A norm check on
each partial vector keeps it within a bounded codebook; the procedure
returns zero if the check fails. All such failures lie in a single event
controlled by the total estimation error, so their risk contribution stays
of order $\varepsilon^2$.

The group schedule also adapts to available workspace. Write
$a=1+\log(1/\varepsilon)$, $h=\tfrac12\log d$, and let $\beta d$ be
the logarithmic state capacity available beyond answer encoding. A fraction
approximately $\theta=\min\{1,(a+\beta)/(a+h)\}$ of the remaining
coordinates is activated at each stage. Groups shrink as the stored part
grows. Their scalar resolutions balance total squared error and stream
duration. Writing $K_{\rm used}$ and $N_{\rm used}$ for its resource use,
the resulting construction satisfies
\begin{align}
\log K_{\rm used}&\le d\log(C/\varepsilon)+\beta d,\label{eq:group-resources}\\
N_{\rm used}&\le C\frac d{\varepsilon^2}
\left(1+\frac{\log d}{1+\log(1/\varepsilon)+\beta}\right)
\nonumber
\end{align}
for mean squared error of order $\varepsilon^2$.
The first resource bound splits the state budget between answer precision
and active workspace. Increasing the latter allows more coordinates to be
processed together, reducing the sample cost in the second bound. Optimizing
subject to $K_{\rm used}\le K$ gives the upper half of
\eqref{eq:main-highdim}. Fine working precision is shared over time, while
finished answers occupy a compact representation.

\section{Baseline protection with increasing confidence}\label{sec:confidence}

We first take $d=1$ and allow $\delta$ to decrease. A candidate estimate
followed by validation on fresh observations achieves high confidence
with a small state space. The validation step introduces a separate
tradeoff between sample size and memory. Set
\begin{equation}\label{eq:LUV}
L=1+\log(1/\delta),\qquad U=1+\log L,\qquad V=1+\log U.
\end{equation}

Fix $\rho\in(0,1/2)$ and let
\[
N_\rho(K,\delta)=\min\left\{n\in\N:\ \exists\widehat w\in
\mathcal A^{\rm str}_{n,K,\delta}(1),\quad
\sup_{|\mu|\le1}\E_\mu R_\mu(\widehat w)\le\rho\right\},
\]
with the minimum of an empty set taken to be infinity.

\begin{theorem}[Sample size and state requirements at fixed risk]
\label{thm:main-confidence}
For every fixed $\rho\in(0,1/2)$ there are positive constants
$c_\rho,C_\rho,D_\rho,\delta_\rho$ such that, whenever
$0<\delta\le\delta_\rho$,
\begin{equation}\label{eq:fixedrisk-confidence}
c_\rho L\max\{1,U/K\}
\le N_\rho(K,\delta)
\le C_\rho L\max\{1,U/K\}.
\end{equation}
The lower bound holds for every $K=2^b$; the upper bound holds when
$K\ge D_\rho V$.
\end{theorem}

The confidence cost is $L\asymp\log(1/\delta)$ samples. Below order $U$
states, validation with finite memory adds a factor $U/K$. At order $U$ states,
sample complexity returns to the ordinary confidence scale. Thus a slowly
growing state space recovers full sample efficiency; matching begins at
the explicit threshold $D_\rho V$.

\begin{corollary}[Exponential protection at a prescribed horizon]
\label{cor:main-exponential}
For each fixed $\rho\in(0,1/2)$ there exists $\gamma_0(\rho)>0$ such that,
for every fixed $0<\gamma\le\gamma_0(\rho)$ and $\delta_n=e^{-\gamma n}$,
the smallest admissible number of states for risk at most $\rho$ satisfies
\[
K_{\min}(n,\rho,\delta_n)=\Theta_{\rho,\gamma}(\log n).
\]
The minimum number of persistent bits is
$\log_2\log n+O_{\rho,\gamma}(1)$.
\end{corollary}

This guarantee concerns the decoded action at the prescribed horizon $n$.
Under the same exponential baseline protection, a fixed terminal alphabet
suffices at fixed risk for sufficiently small $\gamma$
(Proposition~\ref{prop:conf-terminal}). The growing cost of working memory
therefore comes from validating the decision to act. With bounded working
states, the minimax risk instead tends to the value of the zero baseline, $1/2$
(Proposition~\ref{conf:collapse}).

\paragraph{Testing and validation.}
Safety at zero and risk improvement at a separated mean induce a test with
small type I error and fixed positive power. Monotone reduction turns it
into an ordered test with finitely many states. A bound on weighted certificate area,
together with fixed power, yields a short acceptance certificate. Its
probability under the null limits the achievable type I error. For a fixed
Gaussian mean separation and fixed power, the test decision $D$ satisfies
$-\log P_0(D=1)\le CKn/\log n$ for large $n$.
Combining the state restriction with ordinary testing information yields
the lower bound. Appendix~\ref{app:confidence} also gives the joint risk
bound in $n,K,\delta$ (Theorem~\ref{thm:main-confidence-lower}).

The upper bound first forms a pilot estimate at fixed resolution. It then
validates a conservative update using an AND of repeated OR tests on fresh
blocks. The OR tests retain power; their AND suppresses unsafe acceptance.
The public clock schedules these tests, and their work states are reused
across a fixed collection of candidate scales. This asymmetric validation
attains \eqref{eq:fixedrisk-confidence}.

The same validation mechanism applies coordinate by coordinate in any fixed
dimension. The constants then depend on the dimension and target risk,
while the dependence on confidence and total states retains its scalar form.
For dimension $d$, let $N_\rho^{(d)}(K,\delta)$ be the minimum sample size
for a procedure in $\mathcal A^{\rm str}_{n,K,\delta}(d)$ with worst-case
risk at most $\rho$, or infinity if none exists.

\begin{corollary}[Confidence costs in fixed dimension]
\label{cor:main-fixed-d-confidence}
Fix an integer $d\ge1$ and $\rho\in(0,1/2)$. There are positive constants
$c_{d,\rho},C_{d,\rho},D_{d,\rho},\delta_{d,\rho}$ such that, for
$0<\delta\le\delta_{d,\rho}$,
\begin{equation}\label{eq:fixed-d-confidence}
c_{d,\rho}L\max\{1,U/K\}
\le N_\rho^{(d)}(K,\delta)
\le C_{d,\rho}L\max\{1,U/K\}.
\end{equation}
The lower bound holds for every $K=2^b$; the upper bound holds for
$K\ge D_{d,\rho}V$. Consequently, there is $\gamma_0(d,\rho)>0$ such that,
for every fixed $0<\gamma\le\gamma_0(d,\rho)$ and $\delta_n=e^{-\gamma n}$,
the minimum state and bit budgets for risk at most $\rho$ satisfy, as
$n\to\infty$,
\[
K_{\min}^{(d)}(n,\rho,\delta_n)=\Theta_{d,\rho,\gamma}(\log n),
\qquad b_{\min}^{(d)}=\log_2\log n+O_{d,\rho,\gamma}(1).
\]
\end{corollary}

Sequential coordinate blocks reuse one validation workspace, retaining
completed candidate labels. Each coordinate receives failure probability
$\delta/d$; restriction to a coordinate axis gives the matching lower bound.
The scalar confidence law thus extends to every fixed dimension.
Appendix~\ref{app:fixed-d-confidence} gives the proof and resource accounting.

\section{Conclusion}\label{sec:discussion}

The memory required to compute a mean can exceed the memory required to
represent it. At fixed confidence, scalar streaming estimation attains the
rate for terminal encoding. In higher dimensions, statistically optimal sample
size at a fixed sufficiently small risk requires order $d\log d$ working
bits, compared with order $d$ bits for the answer. Increasing confidence
introduces a further cost: at fixed risk in every fixed dimension, failure
probability $e^{-\gamma n}$ for sufficiently small fixed $\gamma>0$ requires
and permits $\log_2\log n+O(1)$ working bits.

The constructions attain these bounds by separating stored estimates from
active computation. Completed estimates occupy compact labels, and the
workspace is reused for modular measurements, coordinate groups, and
validation. The matching lower bounds identify when this workspace is
statistically necessary. Together, the results quantify how a memory budget
controls answer precision, sample efficiency, and confidence.

\section*{Acknowledgments}
This work was supported by the National Natural Science Foundation of China
under Grant 32270689.

\section*{AI use statement}
OpenAI's ChatGPT/Codex was used to assist with mathematical reasoning and
proof checking, reference verification, and manuscript organization and
language editing. The authors are responsible for the mathematical claims,
citations, and final content of this manuscript.

% Metadata checked against publisher pages, author copies, and the cited
% arXiv versions.  The 2022 Acharya et al. entry identifies the version read.
{\raggedright
\setlength{\bibsep}{0pt}
\interlinepenalty=10000

}

\clearpage
\appendix
\setstretch{1.0}
\part*{Appendix: Proofs}
\noindent The proofs follow the construction's dependency structure.
Appendix~\ref{app:scalar} establishes the scalar engine;
Appendix~\ref{app:highdim} combines it with the information bound and
grouped allocation; Appendix~\ref{app:confidence} develops the asymmetric
tests needed when baseline protection increases.
\section{Scalar estimation with a finite state space}
\label{app:scalar}

This appendix proves the scalar construction used in
Theorem~\ref{thm:main-scalar} and in the upper bound for several dimensions.
Its statistical input is a Bernoulli stream; Gaussianity enters only in
the final conversion from a sign probability to a mean.
Throughout this appendix, put
\begin{equation}
 A=16384,\qquad \epsilon_0=\frac1{64},\qquad
 \kappa=\frac1{100},\qquad
 \rho=\frac9{32},\qquad c_\star=\frac9{16},\qquad
 D=\frac{1377}{4096}.
 \label{eq:scalar-constants}
\end{equation}
Write
$\operatorname{dist}_{\mathbb T}(u,v)=\min_{m\in\mathbb Z}|u-v-m|$
for distance on $\mathbb T=\mathbb R/\mathbb Z$.

\begin{lemma}[Scalar estimation subroutine]
\label{lem:scalar-engine}
Let $a=\Phi(-1)$ and $B=D/[a^2\phi(1)^2]$.
For every power of two $M\ge2$, there is a deterministic streaming
estimator $q$ for observations $X_i\sim N(\mu,1)$, $|\mu|\le1$, which
uses at most $2AM$ states and $32AM^2$ observations and satisfies
\begin{equation}
 q\in[-1,1],\qquad
 \sup_{|\mu|\le1}\mathbb E_\mu(q-\mu)^2\le\frac{B}{M^2}.
 \label{eq:scalar-engine}
\end{equation}
The sample schedule is fixed before any observations are read.
\end{lemma}

We first construct an estimator of a Bernoulli probability on the
circle, then prove the lemma by decoding its final state.

\subsection{Modular measurements and recovery from two candidates}
\label{app:scalar-measurement}

For an integer frequency $f\ge1$, a measurement uses
$L=Af^2$ observations $Y_i\sim\operatorname{Bernoulli}(p)$ and the
counter
\begin{equation}
 V=Af,\qquad C_0=0,\qquad
 C_t=(C_{t-1}+Y_t)\bmod V,\qquad z=C_L/V.
 \label{eq:scalar-measurement}
\end{equation}
Only $C_t\in\{0,\ldots,V-1\}$ persists between observations. Since
$L/V=f$, the block output satisfies
\[
 \operatorname{dist}_{\mathbb T}(z,fp)
 \le\left|\frac{\sum_{i=1}^L Y_i}{V}-fp\right|
 =f|\overline Y-p|.
\]
Hoeffding's inequality \citep{hoeffding1963} gives, uniformly over $p\in[0,1]$,
\begin{equation}
 \mathbb P_p\{\operatorname{dist}_{\mathbb T}(z,fp)>\epsilon_0\}
 \le2\exp(-2A\epsilon_0^2)=2e^{-8}<\kappa.
 \label{eq:scalar-measurement-tail}
\end{equation}
Thus $Af$ states suffice to measure the phase $fp\bmod1$ to fixed
accuracy with fixed confidence, using $Af^2$ samples.

The next elementary observation explains how such measurements
recover a probability while the frequencies decrease.

\begin{lemma}[Recovery from a doubled phase]
\label{lem:scalar-recovery}
Let $r\in[0,1)$, $\theta\in\mathbb T$, and
$d=\operatorname{dist}_{\mathbb T}(r,2\theta)$.
For the two candidates
$c_0=r/2$ and $c_1=(r+1)/2$, their distances to $\theta$ are the
unordered pair
\begin{equation}
 \left\{\operatorname{dist}_{\mathbb T}(c_0,\theta),
        \operatorname{dist}_{\mathbb T}(c_1,\theta)\right\}
 =\left\{\frac d2,\frac12-\frac d2\right\}.
 \label{eq:scalar-root-distances}
\end{equation}
If $d\le7/16$ and
$\operatorname{dist}_{\mathbb T}(z,\theta)\le\epsilon_0$, the candidate
nearest to $z$ is the one nearest to $\theta$, and its error is $d/2$.
If $d>7/16$, either candidate has error at most $\rho=9/32$.
\end{lemma}

\begin{proof}
Rotate the circle by $-r/2$, so the candidates become $0$ and $1/2$.
If $q=\operatorname{dist}_{\mathbb T}(0,\theta-r/2)$, their distances
are $q$ and $1/2-q$, while
$d=\min\{2q,1-2q\}$. This proves
\eqref{eq:scalar-root-distances}. When $d\le7/16$, the gap between
the two distances is $1/2-d\ge4\epsilon_0$.
Replacing $\theta$ by $z$ changes this gap by at most $2\epsilon_0$,
so the nearer candidate remains nearer. When $d>7/16$, the larger
distance is $1/2-d/2<1/4+2\epsilon_0=\rho$.
\end{proof}

The recovery case is essential: a mistaken earlier decision need not
invalidate every later measurement. One accurate layer brings an
arbitrary previous error into a region where subsequent accurate
layers halve it.

\subsection{Amplification with four states}
\label{app:scalar-amplification}

\begin{lemma}[Amplifier with four states]
\label{lem:four-state-amplifier}
Suppose independent, identically distributed votes $U_i\in\{0,1\}$
satisfy $\mathbb P(U_i=u_*)\ge99/100$ for some fixed $u_*$.
For $0<\eta<1$, define
\begin{equation}
 \ell=\left\lceil\frac{\log(2/\eta)}{\log99}\right\rceil,
 \qquad
 T=\left\lceil(100/99)^\ell\log(2/\eta)\right\rceil.
 \label{eq:scalar-amplifier-schedule}
\end{equation}
There is a deterministic program with four states that reads
$\ell T$ votes and returns $u_*$ with probability at least $1-\eta$.
\end{lemma}

\begin{proof}
The program stores a current decision $F\in\{0,1\}$ and a flag
$H\in\{0,1\}$. Each of $T$ rounds contains $\ell$ votes.
At the start of a round set $H=1$ and hold $F$ fixed throughout
that round. Whenever a vote equals $F$, set $H=0$.
At the end of the round, flip $F$ if and only if $H=1$.
Thus a decision changes only when every vote in a round opposes it.
The public clock gives the round boundaries; no votes or round
counters are stored. This construction with prescribed runs is related to
the tests with transitions that vary over time of \citet{cover1969}; the following argument
gives the bound on error probability used here.

Assume $u_*=1$ and let $u=\mathbb P(U_i=1)\ge99/100$.
At round boundaries, $F$ is a Markov chain with two states with transition
probabilities
\[
 \alpha=u^\ell\quad(0\longrightarrow1),\qquad
 \beta=(1-u)^\ell\quad(1\longrightarrow0).
\]
Since $\ell\ge1$, $0<\alpha+\beta\le1$.
For $w_t=\mathbb P(F_t=0)$, the recursion
$w_{t+1}=\beta+(1-\alpha-\beta)w_t$ yields, for any initial decision,
\begin{align}
 w_T
 &\le\frac\beta\alpha+\exp(-T\alpha)\notag\\
 &\le99^{-\ell}+\exp\{-T(99/100)^\ell\}
 \le\eta.
 \label{eq:scalar-amplifier-error}
\end{align}
Interchanging labels proves the other case. The only persistent
variables are $(F,H)$, giving four states.
\end{proof}

\subsection{The frequency schedule and its memory cost}
\label{app:scalar-schedule}

Fix $M=2^J$, $J\ge1$, and predetermined target errors
$\eta_1,\ldots,\eta_J\in(0,1)$.
Layer $j$ uses frequency $f_j=M/2^j$ and starts with the label
\[
 r_{j-1}=k/2^{j-1},\qquad 0\le k<2^{j-1},\qquad r_0=0.
\]
Keep this label fixed during the layer. Each fresh block of
$Af_j^2$ samples produces the modular measurement
\eqref{eq:scalar-measurement} and casts one vote for the nearer of
the candidates $r_{j-1}/2$ and $(r_{j-1}+1)/2$; break ties in favor
of the first candidate. Use the amplifier with target error $\eta_j$
and initial decision $F=0$, producing its final decision after
$T_j$ rounds of $\ell_j$ votes each. Set
\begin{equation}
 r_j=\frac{r_{j-1}+F}{2}.
 \label{eq:scalar-label-update}
\end{equation}
The final output is $r_J\in\{0,1/M,\ldots,(M-1)/M\}$.

At every sample time in layer $j$, all persistent information is
\begin{equation}
 (k,F,H,C),\qquad
 0\le k<2^{j-1},\quad F,H\in\{0,1\},\quad 0\le C<Af_j.
 \label{eq:scalar-state-tuple}
\end{equation}
Its number of possible values is
\begin{equation}
 2^{j-1}\cdot4\cdot Af_j=2AM.
 \label{eq:scalar-state-count}
\end{equation}
In particular, the growing label and the shrinking counter exactly
offset each other. A layer changes the interpretation of a common
state set; it does not add a new state set to those of earlier layers.

For completeness, the implementation updates memory after every
sample. An ordinary transition increments $C$ modulo $Af_j$.
At a block boundary it first forms the updated counter, computes
the vote temporarily, and updates $H$. If that boundary also ends
a round, it then applies the possible flip of $F$ and resets the
flag for the next round. At a layer boundary it replaces the label
by
\[
 k_{\mathrm{new}}=k+F2^{j-1},
\]
which implements \eqref{eq:scalar-label-update} exactly, and resets
the work variables. The public sample index determines every block,
round, and layer boundary. These operations compose into one
transition, so neither the temporary vote nor its normalized phase
needs an extra persistent state. The final $M$ labels also fit in
the same state set. After the scheduled computation, the program
can freeze its state and ignore any remaining observations.

The total number of observations is exactly
\begin{equation}
 N(M;\eta_1,\ldots,\eta_J)
 =A\sum_{j=1}^J f_j^2\ell_jT_j.
 \label{eq:scalar-exact-samples}
\end{equation}
For the construction at fixed confidence choose
\begin{equation}
 \eta_j=2^{-3j}/16.
 \label{eq:scalar-layer-errors}
\end{equation}
Writing $t_j=\log(2/\eta_j)=\log32+3j\log2$ gives
$\ell_j\le j+1$, $t_j\le4(j+1)$, and
$T_j\le4(j+1)(100/99)^{j+1}+1$. Consequently,
\begin{align}
 \frac{N(M)}{AM^2}
 &=\sum_{j=1}^J4^{-j}\ell_jT_j\notag\\
 &\le8\sum_{j=1}^\infty\frac{(j+1)^2}{3^j}
       +\sum_{j=1}^\infty\frac{j+1}{4^j}
 =28+\frac79<32.
 \label{eq:scalar-total-samples}
\end{align}
Here we used $(100/99)/4<1/3$ and $100/99<2$.
The increasingly accurate decisions are affordable because the
blocks become geometrically shorter.

\subsection{Error analysis through the last incorrect layer}
\label{app:scalar-error}

\begin{lemma}[Last incorrect layer]
\label{lem:scalar-last-bad}
For the preceding algorithm with any predetermined $\eta_j$, let
$d_0=\operatorname{dist}_{\mathbb T}(0,Mp)$ and
$d_j=\operatorname{dist}_{\mathbb T}(r_j,f_jp)$.
Then
\begin{equation}
 \mathbb E_p d_J^2
 \le\frac{c_\star^2}{M^2}
       +c_\star^2\sum_{j=1}^J\frac{\eta_j}{f_j^2}.
 \label{eq:scalar-last-bad}
\end{equation}
\end{lemma}

\begin{proof}
Call layer $j$ bad if $d_{j-1}\le7/16$ and its amplifier chooses
the candidate farther from the true phase $f_jp\bmod1$.
Let $\mathcal F_{j-1}$ contain all observations read before this
layer. Conditional on $\mathcal F_{j-1}$, the old label is fixed.
If $d_{j-1}\le7/16$, Lemma~\ref{lem:scalar-recovery} and
\eqref{eq:scalar-measurement-tail} imply that each vote chooses the
unique nearer candidate with probability at least $99/100$.
The votes are conditionally independent and identically distributed:
their sample blocks are fresh and disjoint, and their rule depends
on the fixed old label, not on the current decision $F$.
Lemma~\ref{lem:four-state-amplifier} thus gives
\begin{equation}
 \mathbb P_p(B_j\mid\mathcal F_{j-1})\le\eta_j.
 \label{eq:scalar-conditional-error}
\end{equation}
If $d_{j-1}>7/16$, the bad event $B_j$ is empty by definition.
Hence this conditional bound holds regardless of earlier errors.

On $B_j^c$, Lemma~\ref{lem:scalar-recovery} gives $d_j\le\rho$;
if also $d_{j-1}\le7/16$, it gives $d_j=d_{j-1}/2$.
Because $\rho<7/16$, the first good layer recovers a controlled
error and every consecutive good layer thereafter halves it.
If there are no bad layers, therefore,
\[
 d_J\le\rho\,2^{-(J-1)}=c_\star/M.
\]
If $j<J$ is the last bad layer, layer $j+1$ first restores an
error of at most $\rho$ and the remaining layers halve it, giving
\[
 d_J\le\rho\,2^{-(J-j-1)}=c_\star/f_j.
\]
For $j=J$ the same bound follows from $d_J\le1/2\le c_\star/f_J$.
Partition the sample space according to the last bad layer, including
the event that none occurs. The event that $j$ is the last bad
layer is contained in $B_j$ and has probability at most $\eta_j$
by \eqref{eq:scalar-conditional-error}. Summing its bound on squared error proves \eqref{eq:scalar-last-bad}.
\end{proof}

With \eqref{eq:scalar-layer-errors} and $f_J=1$, this yields
\begin{equation}
 \sup_{p\in[0,1]}
 \mathbb E_p\operatorname{dist}_{\mathbb T}(r_J,p)^2
 \le\frac{c_\star^2}{M^2}
       \left(1+\frac1{16}\sum_{j=1}^\infty2^{-j}\right)
 =\frac{D}{M^2}.
 \label{eq:scalar-circular-risk}
\end{equation}
Together with \eqref{eq:scalar-state-count} and
\eqref{eq:scalar-total-samples}, this proves the circular Bernoulli
subroutine with $O(M)$ states, $O(M^2)$ samples, and squared error
$O(M^{-2})$.

\begin{proof}[Proof of Lemma~\ref{lem:scalar-engine}]
Form $Y_i=\mathbf1\{X_i\ge0\}$ as each observation arrives. These
are independent $\operatorname{Bernoulli}(p)$ observations with
$p=\Phi(\mu)\in[a,1-a]$. Run the circular subroutine and decode
its final label $r=r_J$ as
\begin{equation}
 q=\Phi^{-1}\!\left(\Pi_{[a,1-a]}(r)\right).
 \label{eq:scalar-gaussian-decoder}
\end{equation}
For every $p\in[a,1-a]$ and $r\in[0,1)$,
\begin{equation}
 |r-p|\le a^{-1}\operatorname{dist}_{\mathbb T}(r,p).
 \label{eq:scalar-unwrapping}
\end{equation}
Indeed, if $|r-p|\le1/2$, circular and ordinary distances agree.
Otherwise their circular distance is $1-|r-p|\ge a$, whereas
$|r-p|\le1$.
Projection onto $[a,1-a]$ cannot increase distance to $p$, and
the derivative of $\Phi^{-1}$ on that interval is at most
$1/\phi(1)$. Thus
\[
 |q-\mu|\le
 \frac{\operatorname{dist}_{\mathbb T}(r,p)}{a\phi(1)}.
\]
Equation~\eqref{eq:scalar-circular-risk} proves the claimed mean
squared error. The sign transformation is performed within a
single transition and the transformation using the inverse cdf only at
decoding, so neither changes the persistent state count.
\end{proof}

The same construction estimates an arbitrary mean on $[0,1]$.
For $X_i\in[0,1]$, generate an independent uniform $U_i$ and form
$Y_i=\mathbf1\{U_i\le1/4+X_i/2\}$.
Then $\mathbb E Y_i=p=1/4+(\mathbb E X_i)/2\in[1/4,3/4]$.
Decoding by $\widehat m=2\Pi_{[1/4,3/4]}(r_J)-1/2$ and applying
\eqref{eq:scalar-unwrapping} with $a=1/4$ gives
$\mathbb E(\widehat m-\mathbb E X_i)^2\le64D/M^2$ with the
same resources. This reduction uses fresh randomness but no
additional persistent memory.

\subsection{The scalar minimax rate at fixed confidence}
\label{app:scalar-rate}

\begin{proof}[Proof of Theorem~\ref{thm:main-scalar}]
For the upper bound, let
\[
 H=\min\left\{\frac{K}{2A},\sqrt{\frac{n}{32A}}\right\}.
\]
If $H\ge2$, take the largest power of two $M\le H$, so that
$H/2<M\le H$. Lemma~\ref{lem:scalar-engine} fits within both
resource budgets. Apply Lemma~\ref{lem:radial-safety} in one
dimension with the deterministic threshold
$\lambda=\sqrt{B/\delta}/M$. The resulting safe estimator has
risk at most
\[
 \frac{B(1+\delta^{-1})}{M^2}
 \le4B(1+\delta^{-1})
       \max\left\{\frac{4A^2}{K^2},\frac{32A}{n}\right\}
 \le C_\delta(K^{-2}+n^{-1}).
\]
If $H<2$, the zero estimator suffices: its risk is at most $1/2$,
and either $K<4A$ or $n<128A$, so $K^{-2}+n^{-1}$ is bounded
below by an absolute positive constant. The zero estimator is
also available in the first case. Taking the better predetermined
choice gives the asserted upper bound with the truncation at one.

The lower bound holds even for terminal compression without the
safety requirement. For its statistical term, take
$\mu=\pm h$, where $h=1/(2\sqrt n)$. The relative entropy of the
two observation laws is $2nh^2=1/2$, so their total variation
distance is at most $1/2$ by Pinsker's inequality. Any sign test
has average error at least $1/4$. Applying the test
$\mathbf1\{\widehat w>0\}$ shows that the worst-case loss
$\tfrac12(\widehat w-\mu)^2$ is at least $1/(32n)$.

For the memory term, put the uniform prior on $[-1,1]$ and fix all
independent algorithmic randomness, including output randomness.
The resulting decoder has at most $K$ output values. The union of
their intervals of radius $s=1/(2K)$ has length at most one.
At least half the prior mass is therefore at distance at least
$s$ from every output value, even if an encoder were given the
true parameter. Its loss averaged over the prior is at least
$s^2/4=1/(16K^2)$. This bound survives averaging over the
randomness. Combining the two terms yields
\[
 \sup_{|\mu|\le1}\mathbb E_\mu\frac{(\widehat w-\mu)^2}{2}
 \ge\max\{(32n)^{-1},(16K^2)^{-1}\}
 \ge\frac1{64}\min\{1,n^{-1}+K^{-2}\}.
\]
Finally, a terminal encoder can simulate every streaming program
and transmit its final state. Its minimax risk is at most the
streaming risk, so the same lower bound and the streaming upper
bound also establish the scalar rate for terminal compression.
\end{proof}

\section{Working memory in several dimensions}\label{app:highdim}

Throughout this appendix, $X_t\sim N_d(\mu,I_d)$ independently, $\mu\in B_2^d$, and
$R_\mu(w)=\|w-\mu\|_2^2/2$.  Logarithms are natural.  All state counts refer to
the complete persistent state; the deterministic time index is public.

\subsection{Safety and the terminal benchmark}\label{app:hd-terminal}

The following observation allows safety at fixed confidence to be imposed after
constructing an estimator with a mean squared error guarantee.

\begin{lemma}[Radial safety transformation]\label{lem:radial-safety}
Suppose that $q$ is an estimator satisfying
$\sup_{\mu\in B_2^d}\mathbb E_\mu\|q-\mu\|_2^2\le r$.
For $0<\delta<1$, set $\lambda=\sqrt{r/\delta}$ and
\[
 w=\begin{cases}
 0,&\|q\|_2\le\lambda,\\
 (1-\lambda/\|q\|_2)q,&\|q\|_2>\lambda.
 \end{cases}
\]
Then $w$ is uniformly $\delta$-safe and
$\sup_\mu\mathbb E_\mu R_\mu(w)\le(1+\delta^{-1})r$.
If $q$ is decoded from a finite state, this transformation requires no
additional states.
\end{lemma}

\begin{proof}
On $\{\|q-\mu\|_2\le\lambda\}$, write $t=\|q\|_2$.
When $t>\lambda$,
\[
 \langle q,\mu\rangle\ge t(t-\lambda),\qquad
 \langle w,\mu\rangle\ge(t-\lambda)^2=\|w\|_2^2.
\]
Consequently $R_\mu(w)-R_\mu(0)=\|w\|_2^2/2-\langle w,\mu\rangle\le0$.
The same conclusion is immediate when $w=0$.
Markov's inequality bounds the probability of the complementary event by
$\delta$.  Moreover, $\|w-q\|_2\le\lambda$, so
\[
 R_\mu(w)\le\|q-\mu\|_2^2+\lambda^2.
\]
Taking expectations proves the claim.  If $r=0$, use $w=q=\mu$ almost surely.
\end{proof}

\paragraph{Lower bounds common to both memory models.}
Any estimator, even without a memory or safety restriction, satisfies
\begin{equation}\label{eq:hd-stat-lower}
 \sup_{\mu\in B_2^d}\mathbb E_\mu R_\mu(\widehat w)
 \ge\frac1{32}\min\{1,d/n\}.
\end{equation}
Indeed, let $a=\min\{d^{-1/2},(2\sqrt n)^{-1}\}$ and consider
$\mu_\theta=a\theta$ for $\theta\in\{-1,1\}^d$.
Two parameters differing in one sign induce sample laws with relative
entropy $2na^2\le1/2$ and hence total variation at most $1/2$.
For each coordinate, the average error probability of the test
$\mathbf1\{\widehat w_i>0\}$ under the uniform sign prior is at least $1/4$.
A sign error incurs squared error at least $a^2$.  The Bayes risk is therefore
at least $da^2/8\ge\min\{1,d/n\}/32$, proving
\eqref{eq:hd-stat-lower}.

Any procedure whose final decoder receives at most $K$ labels also satisfies
\begin{equation}\label{eq:hd-quant-lower}
 \sup_{\mu\in B_2^d}\mathbb E_\mu R_\mu(\widehat w)
 \ge\frac1{16}K^{-2/d}.
\end{equation}
To see this, put the uniform volume prior on $B_2^d$ and fix the algorithm's
random variables whose laws are independent of the parameter, including its decoder randomization.
There are then at most $K$ possible decoded points.  Their balls of radius
$u=(2K)^{-1/d}$ cover at most half the volume of $B_2^d$.
At least half the prior parameters are at distance at least $u$ from every
decoded point, even if the encoder is given the true parameter.
The prior risk is at least $u^2/4\ge K^{-2/d}/16$ for each fixed random seed.
Averaging over the seed proves the randomized version.  Combining the two
bounds gives, for either memory model,
\begin{equation}\label{eq:hd-common-lower}
 \sup_\mu\mathbb E_\mu R_\mu(\widehat w)
 \ge\frac1{64}\min\{1,d/n+K^{-2/d}\}.
\end{equation}

\paragraph{Proof of the terminal upper bound in Theorem~\ref{thm:main-terminal}.}
A maximal $u$-separated subset of $B_2^d$ is a $u$-net containing at most
$(1+2/u)^d$ points: the disjoint balls of radius $u/2$ about its points lie in
the ball of radius $1+u/2$.
If $v=K^{1/d}\ge3$, take $u=3/v$, so that the net has at most $K$ points.
Project $\overline X$ onto $B_2^d$ and encode its nearest net point $q$.
Projection does not increase distance to $\mu$, and therefore
\[
 \mathbb E_\mu\|q-\mu\|_2^2
 \le 2d/n+2u^2\le18(d/n+K^{-2/d}).
\]
Lemma~\ref{lem:radial-safety} gives the safe upper bound, using the same labels.
If $v<3$, the quantization term exceeds $1/9$ and the zero estimator suffices.
Choosing the zero estimator also whenever the displayed rate exceeds a
constant proves the bound with the outer minimum with $1$.
Together with \eqref{eq:hd-common-lower}, this proves
Theorem~\ref{thm:main-terminal}.

\subsection{A Fisher information bound for joint states}\label{app:hd-information}

Fix $\mu\in\mathbb R^d$ and write
\[
 p_{t,s}(\mu)=\mathbb P_\mu(S_t=s),\quad
 a_t(s)=\nabla\log p_{t,s}(\mu),\quad A_t=a_t(S_t),\quad
 V_t=\mathbb E_\mu\|A_t\|_2^2.
\]
Only reachable states are included.  Thus $V_t$ is the trace of the Fisher
information matrix $\mathcal J_t(\mu)$ of the state $S_t$.
The random vector $A_t$ is a proof device, not additional algorithmic memory.

State probabilities are smooth: after integrating out the randomization,
each is an integral of a bounded measurable function against the Gaussian
sample density.  Differentiation under the integral is valid on every compact
parameter set.  Positivity of Gaussian densities also implies that a state
is either unreachable for every parameter or has positive probability for
every parameter.  Put $Z_t=X_t-\mu$.  Differentiating the transition identity
for $p_{t,s}$ gives the recursion for the conditional score
\begin{equation}\label{eq:hd-score-recursion}
 A_t=\mathbb E_\mu[A_{t-1}+Z_t\mid S_t],\qquad A_0=0.
\end{equation}
For completeness, if $Q_t(s,x;s')$ is the transition probability after
integrating out fresh randomness, then
\[
 p_{t,s'}=\sum_s p_{t-1,s}\int Q_t(s,x;s')\phi_d(x-\mu)\,dx.
\]
Its gradient is the same integral with an additional factor
$a_{t-1}(s)+x-\mu$; division by $p_{t,s'}$ proves
\eqref{eq:hd-score-recursion}.
Let $g_t=V_t-V_{t-1}$.  Since $Z_t$ is independent of $S_{t-1}$ and has
covariance $I_d$, orthogonality of conditional expectation yields
\begin{equation}\label{eq:hd-score-loss}
 \mathbb E_\mu\|A_{t-1}+Z_t-A_t\|_2^2=d-g_t.
\end{equation}
In particular, $g_t\le d$, but $g_t$ need not be nonnegative.

\begin{lemma}[Entropy loss after Gaussian smoothing]\label{lem:hd-smoothing}
Let $A,Y\in\mathbb R^d$ satisfy $\mathbb E[Y\mid A]=A$, where $A$ has finite
support and $\mathbb E\|Y\|_2^2<\infty$.
If $G\sim N_d(0,I_d)$ is independent of $(A,Y)$, then, for $\sigma>0$,
\begin{equation}\label{eq:hd-smoothing}
 \mathsf h(Y+\sqrt\sigma G)-\mathsf h(A+\sqrt\sigma G)
 \le\frac{\mathbb E\|Y-A\|_2^2}{2\sigma}.
\end{equation}
Here $\mathsf h$ denotes differential entropy; both entropies are finite.
\end{lemma}

\begin{proof}
Let $p$ be the finite Gaussian mixture density of $A+\sqrt\sigma G$ and
$f=-\log p$.  Direct differentiation gives
\[
 \nabla^2 f(x)=\frac{I_d}{\sigma}
 -\frac{\operatorname{Cov}(A\mid A+\sqrt\sigma G=x)}{\sigma^2}
 \preceq\frac{I_d}{\sigma}.
\]
Consequently
$f(x+u)\le f(x)+\langle\nabla f(x),u\rangle+\|u\|_2^2/(2\sigma)$.
Use $x=A+\sqrt\sigma G$ and $u=Y-A$.
The linear term has zero expectation because
$\mathbb E[Y-A\mid A,G]=0$, whence
\[
 \mathbb E f(Y+\sqrt\sigma G)
 \le\mathsf h(A+\sqrt\sigma G)
      +\mathbb E\|Y-A\|_2^2/(2\sigma).
\]
Relative entropy nonnegativity bounds
$\mathsf h(Y+\sqrt\sigma G)$ by the cross entropy on the left.
These operations are integrable: one positive mixture component bounds
$f(x)$ above by a constant plus a quadratic function, $p$ is bounded above,
and $\|\nabla f(x)\|_2\le(\|x\|_2+\max\|A\|_2)/\sigma$.
The Gaussian convolution of $Y$ has a bounded density and finite second
moment, which also guarantees its finite entropy.
\end{proof}

\begin{lemma}[Fisher information retained by a finite state]\label{lem:hd-fisher}
For every $n,d,K\ge1$ and every randomized algorithm with transitions that vary over time whose
complete state has at most $K$ values,
\begin{equation}\label{eq:hd-fisher-bound}
 \operatorname{tr}\mathcal J_n(\mu)
 \le\frac{2n(2\log K+d)}{\log(n+2)}
 \le\frac{4n(d+\log K)}{\log(n+2)}
 \qquad(\mu\in\mathbb R^d).
\end{equation}
\end{lemma}

\begin{proof}
For a vector $A$ taking finitely many values, define
\[
 \mathcal I_\sigma(A)=\mathsf h(A+\sqrt\sigma G)
                         -\mathsf h(\sqrt\sigma G).
\]
This is $I(A;A+\sqrt\sigma G)$, so
$0\le\mathcal I_\sigma(A_t)\le H_\mu(S_t)\le\log K$.
Apply Lemma~\ref{lem:hd-smoothing} to $A=A_t$ and
$Y=A_{t-1}+Z_t$. The hypothesis on the conditional mean holds because
$\mathbb E[Y\mid A_t]=\mathbb E[\mathbb E[Y\mid S_t]\mid A_t]=A_t$.
Equations~\eqref{eq:hd-score-recursion}--\eqref{eq:hd-score-loss}
and the Gaussian entropy formula give
\begin{equation}\label{eq:hd-entropy-recursion}
 \mathcal I_\sigma(A_t)\ge\mathcal I_{\sigma+1}(A_{t-1})
 +\frac d2\log(1+1/\sigma)-\frac{d-g_t}{2\sigma}.
\end{equation}
For a fixed prefix length $m$, set $\sigma=m-t+1$ and sum over $t=1,\ldots,m$.
The information terms telescope, $A_0=0$, and
$\sum_{j=1}^m\log(1+1/j)=\log(m+1)$.  Thus
\begin{equation}\label{eq:hd-harmonic}
 \sum_{t=1}^m\frac{g_t}{m-t+1}
 \le2H_\mu(S_m)+d\{H_m-\log(m+1)\}
 \le2\log K+d,
\end{equation}
where $H_m=\sum_{j=1}^m1/j$ and $H_m-\log(m+1)\le1$.

For the proof only, extend the algorithm to time $2n$ by freezing its state
after time $n$ and ignoring subsequent observations.  Then $g_t=0$ for $t>n$.
Sum \eqref{eq:hd-harmonic} over $m=1,\ldots,2n$ to obtain
\[
 2n(2\log K+d)\ge\sum_{t=1}^n g_t H_{2n-t+1}.
\]
Although the increments $g_t$ can be negative, summation by parts and
$V_t\ge0$ give
\[
 \sum_{t=1}^n g_t H_{2n-t+1}
 =V_nH_{n+1}+\sum_{t=1}^{n-1}\frac{V_t}{2n-t+1}
 \ge V_n\log(n+2).
\]
This proves \eqref{eq:hd-fisher-bound}.  No coordinatewise factorization of
the state, or restriction on the computation within a transition, was used.
\end{proof}

\begin{lemma}[Risk bound from Fisher information]\label{lem:hd-vantrees}
Define the prior density on
$[-d^{-1/2},d^{-1/2}]^d\subset B_2^d$ by
\[
 \pi_d(\mu)=d^{d/2}\prod_{i=1}^d
 \cos^2(\pi\sqrt d\,\mu_i/2).
\]
Every decoder of $S_n$ satisfies
\begin{equation}\label{eq:hd-vantrees}
 \sup_{\mu\in B_2^d}\mathbb E_\mu R_\mu(\widehat w)
 \ge\frac{d^2}{2\{\mathbb E_{\pi_d}\operatorname{tr}\mathcal J_n(\mu)
                         +\pi^2d^2\}}.
\end{equation}
\end{lemma}

\begin{proof}
Projection onto the unit ball followed by averaging over independent decoder
randomness cannot increase risk.  It therefore suffices to consider fixed
decoding points $q_s\in B_2^d$.
The prior has total Fisher information $\pi^2d^2$: each coordinate
contributes
\[
 \pi^2d\int_{-1}^1\sin^2(\pi u/2)\,du=\pi^2d.
\]
Under the joint law $\pi_d(\mu)p_{n,s}(\mu)$ put
$T=\nabla\log\{\pi_d(\mu)p_{n,S_n}(\mu)\}$.
The cross term between the prior and state scores is zero because
$\sum_s\nabla p_{n,s}=0$.  Hence
\[
 \mathbb E\|T\|_2^2
 =\pi^2d^2+\mathbb E_{\pi_d}\operatorname{tr}\mathcal J_n(\mu).
\]
Integration by parts in each coordinate gives
$\mathbb E[(q_{S_n,i}-\mu_i)T_i]=1$; the boundary term vanishes because
$\pi_d$ is zero on the boundary faces.  Summing and applying
Cauchy--Schwarz yields
\[
 d^2\le\mathbb E\|q_{S_n}-\mu\|_2^2\,\mathbb E\|T\|_2^2.
\]
All quantities are integrable: the Gaussian score identity
$a_n(S_n)=\mathbb E[\sum_{t=1}^n(X_t-\mu)\mid S_n]$ implies
$\operatorname{tr}\mathcal J_n(\mu)\le nd$.
The worst-case risk is at least the prior average, proving the lemma.
\end{proof}

\paragraph{Lower bound in Theorem~\ref{thm:main-highdim}.}
Set $B=\log K$, $a=d/n$, $z=K^{-2/d}$, and
\[
 T_n=\frac{d^2\log(n+2)}{n(d+B)},\qquad
 T_d=\frac{d^2\log(ed)}{n(d+B)}.
\]
Lemmas~\ref{lem:hd-fisher} and~\ref{lem:hd-vantrees} give a risk lower bound
\[
 \frac{T_n}{8+2\pi^2T_n}\ge\frac1{64}\min\{1,T_n\}.
\]
Combine this with \eqref{eq:hd-common-lower} and use
$\max\{\min(1,x),\min(1,y)\}\ge\tfrac12\min(1,x+y)$ to obtain
\[
 \sup_\mu\mathbb E_\mu R_\mu(\widehat w)
 \ge\frac1{128}\min\{1,a+z+T_n\}.
\]
If $n\ge d$, then $\log(n+2)\ge\frac12\log(ed)$, and thus $T_d\le2T_n$.
If $n<d$, the statistical lower bound already gives a positive constant.
In both cases the desired lower bound follows with constant $1/256$.
The argument did not use safety.

\subsection{A grouped construction with reusable working memory}\label{app:hd-upper}

We use the scalar engine of Lemma~\ref{lem:scalar-engine} in its Gaussian
form: for every power of two $M\ge2$, it estimates any $\mu_i\in[-1,1]$
using at most $2AM$ states and $32AM^2$ observations, with output $Q_i$
satisfying
\begin{equation}\label{eq:hd-scalar-engine}
 \mathbb E(Q_i-\mu_i)^2\le B_0/M^2,
 \qquad A=16384,
\end{equation}
where $B_0$ is absolute.  Identical scalar programs can run in parallel on
different coordinates of the same vector observations.

\begin{proposition}[Groups, resolution, and persistent states]\label{prop:hd-groups}
There are absolute constants $C_s\ge e$ and $C_n\ge1$ such that, for every
$d\ge1$, $0<\varepsilon\le1$, and $\beta\ge0$, a deterministic streaming
estimator $q\in B_2^d$ has
\begin{align}
 \sup_{\mu\in B_2^d}\mathbb E_\mu\|q-\mu\|_2^2
 &\le\tfrac32\varepsilon^2,\label{eq:hd-group-error}\\
 \text{number of states}&\le e^{\beta d}(C_s/\varepsilon)^d,
       \label{eq:hd-group-states}\\
 \text{number of vector observations}&\le
 C_n\frac d{\varepsilon^2}
 \max\left\{1,
 \frac{1+\log(1/\varepsilon)+\tfrac12\log d}
      {1+\log(1/\varepsilon)+\beta}\right\}.
       \label{eq:hd-group-samples}
\end{align}
Its schedule is determined before the observations arrive.
\end{proposition}

\begin{proof}
Put
\[
 a=1+\log(1/\varepsilon),\quad h=\tfrac12\log d,\quad
 \vartheta=\min\{1,(a+\beta)/(a+h)\},\quad
 H=\max\{2,\sqrt{8B_0}\}.
\]
Starting with $r_1=d$ unprocessed coordinates, form successive groups of sizes
\[
 g_j=\max\{1,\lfloor\vartheta r_j\rfloor\},\qquad
 r_{j+1}=r_j-g_j,
\]
until no coordinates remain, and let $S=\sum_j\sqrt{g_j}$.
For group $j$ choose the smallest power of two satisfying
\begin{equation}\label{eq:hd-group-resolution}
 M_j\ge H\sqrt{S\sqrt{g_j}}/\varepsilon.
\end{equation}
Use a fresh consecutive block of vector observations for each group and run
its $g_j$ scalar engines in parallel.  Observations of inactive coordinates
are discarded.  After a group finishes, round its outputs to the grid
$\Delta\mathbb Z$, where $\Delta=\varepsilon/\sqrt d$, and retain only the
rounded coordinates.  If the completed partial vector has norm greater
than $2$, enter a permanent abort state and finally output zero.
Otherwise, after all groups finish, project the completed rounded vector
onto $B_2^d$.

\emph{Bounds on group sizes.}
Every group satisfies $g_j\ge\vartheta r_j/2$.  Therefore
\[
 \sqrt{g_j}
 =\frac{\sqrt{r_j}+\sqrt{r_{j+1}}}{\sqrt{g_j}}
       (\sqrt{r_j}-\sqrt{r_{j+1}})
 \le2\sqrt{2/\vartheta}(\sqrt{r_j}-\sqrt{r_{j+1}}).
\]
Summation gives $S^2\le8d/\vartheta$.
Also $\vartheta d\ge1$, since $a\ge1$ and $(d-1)a\ge h$.
Thus $g_j\le\vartheta d$ even when the maximum defining $g_j$ equals $1$,
and $S\sqrt{g_j}\le2\sqrt2d<3d$.
In particular, the dyadic resolutions satisfy
\begin{equation}\label{eq:hd-group-resolution-upper}
 M_j<2H\sqrt{S\sqrt{g_j}}/\varepsilon
       \le2H\sqrt{3d}/\varepsilon.
\end{equation}

\emph{Error.}
For analysis, define all scalar and rounded outputs $Q_i,Z_i$ on their
prespecified sample blocks even if the actual algorithm has aborted.
Equation~\eqref{eq:hd-group-resolution} gives
\[
 \sum_{i=1}^d\mathbb E(Q_i-\mu_i)^2
 \le B_0\sum_j\frac{g_j}{M_j^2}
 \le\frac{B_0\varepsilon^2}{H^2S}\sum_j\sqrt{g_j}
 \le\varepsilon^2/8.
\]
Since $|Z_i-Q_i|\le\Delta/2$,
$\mathbb E\|Z-\mu\|_2^2\le3\varepsilon^2/4$.
If a completed partial vector has norm greater than $2$, its distance from
the corresponding part of $\mu$ exceeds $1$.  The abort event is therefore
contained in $\{\|Z-\mu\|_2>1\}$ and has probability at most
$3\varepsilon^2/4$.  Projection cannot increase error on the nonabort event,
and the squared error of zero is at most $1$.  This proves
\eqref{eq:hd-group-error} without requiring independence between errors.

\emph{State count.}
The number $P_k$ of grid points in the radius-$2$ ball in $k$ completed
coordinates obeys
\begin{equation}\label{eq:hd-grid-count}
 P_k\le\left(\frac{C_g}{\varepsilon}\sqrt{\frac dk}\right)^k,
 \qquad C_g=\tfrac52\sqrt{2\pi e},\qquad P_0=1.
\end{equation}
Indeed, disjoint cubes of side $\Delta$ centered at these grid points fit
inside the radius-$5/2$ ball; its volume is at most
$(\tfrac52\sqrt{2\pi e/k})^k$.
The volume bound for the unit ball follows directly from
\[
 \operatorname{vol}(B_2^k)
 \le e^{k/2}\int_{\mathbb R^k}e^{-k\|x\|_2^2/2}\,dx
 =(2\pi e/k)^{k/2}.
\]

Suppose $k=d-r$ coordinates are complete and the current group has size
$g=\max\{1,\lfloor\vartheta r\rfloor\}$.
The joint state consists of the partial grid label and the $g$ scalar
working states.  Put
$C_* =\max\{C_g,4AH\sqrt3\}$ and $q_0=\log(1/\varepsilon)=a-1$.
By \eqref{eq:hd-group-resolution-upper} and \eqref{eq:hd-grid-count}, the
logarithm of the number of nonabort states is at most
\begin{align}
 &(k+g)(\log C_*+q_0)+\tfrac k2\log(d/k)+gh\notag\\
 &\quad\le d(\log C_*+q_0)+\tfrac k2\log(d/k)
                     +\{gh-(r-g)q_0\}.\label{eq:hd-state-account}
\end{align}
As usual, $k\log(d/k)=0$ at $k=0$.
The final brace is at most $d(\beta+1)$.  To verify this, if $\beta<h$ and
$\lfloor\vartheta r\rfloor\ge1$, then
\[
 g(q_0+h)-rq_0
 \le r\{\vartheta(q_0+h)-q_0\}
 =r(\beta+1-\vartheta)\le d(\beta+1).
\]
If $\beta\ge h$, then $\vartheta=1$ and the brace is $rh\le d\beta$.
In the remaining case $g=1$, it is at most $h\le d$.
Since $k\log(d/k)\le d/e$, equation~\eqref{eq:hd-state-account} bounds the
number of nonabort states by
\[
 e^{\beta d}\{e^{1+1/(2e)}C_*/\varepsilon\}^d.
\]
One extra abort state is absorbed by taking
$C_s\ge2e^{1+1/(2e)}C_*$.
At a group boundary the working states are replaced by the rounded label
in the same transition; the old and new encodings are not retained together.
All indices for groups and positions within groups follow the public schedule.  Different
stages reuse the same state alphabet, so their state counts are maximized,
not added.  The terminal grid state also satisfies the same bound.

\emph{Samples.}
All coordinates in a group read the same vector observations.  Thus the
number of vector observations is at most
\[
 \sum_j32AM_j^2
 \le\frac{128AH^2S}{\varepsilon^2}\sum_j\sqrt{g_j}
 \le\frac{1024AH^2d}{\vartheta\varepsilon^2}.
\]
Taking $C_n\ge1024AH^2$ proves \eqref{eq:hd-group-samples}.
\end{proof}

\paragraph{Upper bound in Theorem~\ref{thm:main-highdim}.}
Write $u=(\log K)/d$ and
\[
 E=K^{-2/d}+\frac dn\left(1+\frac{\log(ed)}{1+u}\right).
\]
Use the constants $C_s\ge e,C_n\ge1$ of Proposition~\ref{prop:hd-groups}.
If $u\le2\log C_s$, then $E\ge C_s^{-4}$ and the zero estimator gives the
desired bound after an absolute adjustment of the constant.
Otherwise fix $C_0\ge\max\{C_s^2,C_n,1\}$ and set $r=C_0E$.
The zero estimator also suffices when $r\ge1$.
When $r<1$, choose
\[
 \varepsilon=\sqrt r,\qquad \beta=u-\log(C_s/\varepsilon).
\]
Because $r\ge C_s^2e^{-2u}$, we have $\beta\ge0$ and
$e^{\beta d}(C_s/\varepsilon)^d=K$.
Moreover, for $a=1+\log(1/\varepsilon)$ and $h=\frac12\log d$,
\[
 a+\beta=1+u-\log C_s\ge(1+u)/2,\qquad
 \max\{1,(a+h)/(a+\beta)\}
 \le1+\frac{h}{a+\beta}
 \le1+\frac{\log(ed)}{1+u}.
\]
The sample bound in Proposition~\ref{prop:hd-groups} is consequently at most
$C_nd\{1+\log(ed)/(1+u)\}/r\le n$.
Freeze the state if the computation finishes before time $n$.
Lemma~\ref{lem:radial-safety} gives risk at most
$\tfrac32(1+\delta^{-1})r\le C_\delta E$.
Choosing zero when $E\ge1$ establishes the asserted outer minimum.

\subsection{Consequences for representation and sample efficiency}\label{app:hd-consequences}

\paragraph{Proof of Corollary~\ref{cor:main-separation}.}
Let $\ell=\log d$, $n_d=\lceil d\ell^2\rceil$, and
$b_d=\lceil d\log_2\ell\rceil$.  Integer rounding gives
\[
 d/n_d\asymp\ell^{-2},\qquad
 K_d^{-2/d}\asymp\ell^{-2},\qquad
 (\log K_d)/d=\log\ell+O(1/d).
\]
The terminal risk is therefore $\Theta_\delta(\ell^{-2})$.
The additional streaming term equals
\[
 \frac{d}{n_d}\frac{\log(ed)}{1+(\log K_d)/d}
 \asymp\frac1{\ell\log\ell},
\]
which dominates $\ell^{-2}$ and still converges to zero.
The claimed risk ratio follows from Theorem~\ref{thm:main-highdim}.

\paragraph{Proof of Corollary~\ref{cor:main-memory}.}
The lower bound in Theorem~\ref{thm:main-highdim} applies without safety.
If the worst-case risk is at most $0<r\le256^{-2}$, its rate expression is
at most $256r<1$.  In particular,
\[
 \log K\ge\frac d2\log\frac1{256r}
              \ge\frac d4\log\frac1r.
\]
If also $d\le n\le Qd/r$, the term for working memory implies
\[
 \log K\ge\frac{d^2\log(ed)}{256rn}-d
              \ge\frac{d\log(ed)}{256Q}-d.
\]
For $\log(ed)\ge512Q$, this yields
$b\ge(d/\log2)\max\{\frac14\log(1/r),\log(ed)/(512Q)\}$.
For sufficiency enlarge the constant in the upper bound to $C_\delta\ge1$ and take
\[
 n\ge8C_\delta d/r,\qquad
 \log K\ge d\{\log(ed)+\tfrac12\log(4C_\delta/r)\}.
\]
The quantization term is then at most $r/(4C_\delta)$ and the bracket
in the sampling term is at most $2$.  Theorem~\ref{thm:main-highdim}
gives risk at most $r$, with
$n=O_\delta(d/r)$ and
$b=O_\delta(d\max\{\log(1/r),\log(ed)\})$.
Rounding the sample size and bit budget up to integers changes neither order.

\section{Reliability and working memory}\label{app:confidence}

Throughout this appendix, $L=1+\ln(1/\delta)$, $U=1+\ln L$, and
$V=1+\ln U$. The testing arguments allow randomized initial states,
randomized transitions, and randomized terminal decisions. All constants
denoted by $C$ are numerical unless their dependence is indicated.

\subsection{Reduction to monotone tests with finite memory}

\begin{lemma}[Monotone reduction]\label{conf:monotone}
Let $P_0=N(u_0,1)$ and $P_1=N(u_1,1)$, where $u_0<u_1$.
For every $K$-state test of $n$ independent observations and every
$\lambda>0$, there is a deterministic $K$-state test with no larger
weighted error
\[
 P_1(D=0)+\lambda P_0(D=1).
\]
Its effective states at each time can be ordered so that every transition
is nondecreasing in both the previous state and the current observation,
and the terminal decision is nondecreasing in the final state.
Consequently, if the original test has power at least $c>0$ and false
positive probability $\xi\in(0,1)$, the monotone test can be chosen with
power at least $c/2$ and false positive probability at most $2\xi/c$.
The same conclusions hold for independent, nonidentically distributed
observations whose individual likelihood ratios are
nondecreasing.
\end{lemma}

\begin{proof}
First fix an initial state whose weighted error does not exceed the
average over the initial randomization. Modify the transitions in
chronological order. Suppose that the preceding transitions have already
been fixed and their effective states are ordered by the prefix
likelihood ratios
\[
 \Lambda(s)=\frac{P_1(S_{t-1}=s)}{P_0(S_{t-1}=s)}.
\]
Both probabilities are positive because the two experiments are
equivalent. For each possible destination $j$, let $a_j$ and $b_j$ be,
respectively, the type II and type I error probabilities of the current
suffix program started from $j$. The suffix may still be randomized.
Writing $f_i$ for the current observation density, the contribution of
choosing destination $j$ at $(s,x)$ is proportional to
\[
 z(s,x)a_j+b_j,\qquad
 z(s,x)=\lambda^{-1}\Lambda(s)\frac{f_1(x)}{f_0(x)}.
\]
A randomized transition is a convex combination of these quantities.
Replace it by a pointwise minimizing destination.

The lower envelope of finitely many affine functions $a_jz+b_j$ admits
an ordered partition of $(0,\infty)$ into intervals on which a minimizing
index is constant. Delete unnecessary functions and resolve each tie
consistently. Ordering the destinations from left to right makes the
selected index $J(z)$ nondecreasing. Thus $J(z(s,x))$ is nondecreasing in
both $s$ and $x$. Under $P_0$, let $Q$ denote the joint law of $(s,x)$.
For a destination interval $I_j$, its new prefix likelihood ratio is
\[
 \frac{P_1(J(z)=j)}{P_0(J(z)=j)}
   =\lambda\mathbb E_Q[z\mid z\in I_j].
\]
The conditional means over distinct ordered intervals of positive
probability are strictly increasing. Indeed, independent draws from two
such intervals are ordered almost surely, and equality of their means
would force both draws to equal the same constant, contradicting the
disjoint partition. Hence the new state ordering also agrees with its
prefix likelihood ratios. Relabel the suffix entrance states, freeze
this transition, and continue. Subsequent modifications cannot change
the probabilities of an already fixed prefix. No step increases the
number of states or the weighted error.

At the final time choose $D=\mathbf1\{\Lambda(S_n)\ge\lambda\}$.
This also minimizes the weighted error and is nondecreasing. Gaussian
likelihood ratios are strictly increasing in $x$, so the induction
applies. It uses only independence and the current monotone likelihood
ratio, proving the nonidentically distributed version as well.
Finally, set $\lambda=c/(2\xi)$. The original weighted error is at most
$1-c/2$, so each of the two nonnegative terms for the new test is at most
this value. Its power is therefore at least $c/2$, and its false positive
probability is at most $(2-c)\xi/c\le2\xi/c$.
\end{proof}

\subsection{A weighted certificate bound}

For a nonconstant monotone function $f:\{0,1\}^n\to\{0,1\}$ and positive
weights $w_i$, a one-certificate is a set of coordinates whose being one
forces $f=1$; a zero-certificate is defined analogously. Let $u(f)$ and
$v(f)$ be the minimum total weights of these certificates.

\begin{lemma}[Certificate area and disjoint certificates]
\label{conf:certificates}
Suppose $f$ is computed by a deterministic monotone read-once program
with at most $K$ states at every time, and put $S=\sum_iw_i^2$.
Then
\begin{equation}\label{conf:area}
 u(f)v(f)\le KS.
\end{equation}
Moreover, writing $u=u(f)$, there are at least $u^2/(4KS)$ pairwise
disjoint zero-certificates, each of weight at most $2KS/u$.
If $Q$ is a product distribution with $Q(X_i=0)=e^{-w_i}$, then
\begin{equation}\label{conf:power-certificate}
 Q(f=1)\le
 \exp\left\{-\frac{u^2}{4KS}\exp\left(-\frac{2KS}{u}\right)\right\}.
\end{equation}
If $P$ is a product distribution with $P(X_i=1)=e^{-w_i}$, then
$P(f=1)\ge e^{-u}$.
\end{lemma}

\begin{proof}
Associate to each nonconstant residual function $g$ the rectangle
$[0,u(g)]\times[0,v(g)]$; constant residual functions have empty
rectangles. For the two children $g_0\le g_1$ at coordinate $i$, use the
conventions $(u(0),v(0))=(\infty,0)$ and
$(u(1),v(1))=(0,\infty)$. Monotonicity gives
\[
 u(g)=\min\{u(g_0),w_i+u(g_1)\},\qquad
 v(g)=\min\{w_i+v(g_0),v(g_1)\}.
\]
Apart from the coordinate axes, any point of the parent rectangle not
covered by either child rectangle belongs to
\[
 (u(g_1),u(g_1)+w_i]\times(v(g_0),v(g_0)+w_i].
\]
Its area is at most $w_i^2$. This includes constant children, since
$u(g_1)$ and $v(g_0)$ remain finite for a nonconstant parent. Taking the
union of rectangles in each layer, its area can increase by at most
$Kw_i^2$ when the layer is traversed backward. The terminal union is
empty. Summing proves \eqref{conf:area}.

For the disjoint certificates, let $I$ be the union of those already
selected. While its weight is below $u/2$, fix all coordinates in $I$ to
one. The restricted function $g$ is nonconstant and has
$u(g)\ge u-w(I)>u/2$. It is still computed by a monotone program with at
most $K$ states: simply compose the transitions for the fixed coordinates.
By \eqref{conf:area}, it has a zero-certificate of weight at most
$2KS/u$, disjoint from $I$. It is also a zero-certificate for $f$, since
lowering the fixed coordinates cannot increase its output. Repeat until
$w(I)\ge u/2$. Each new certificate is nonempty, so the procedure
terminates and selects at least $u^2/(4KS)$ certificates.

Under $Q$, each selected certificate is all zero with probability at
least $\exp(-2KS/u)$, and these events are independent. If any occurs,
$f=0$. Their product bound gives \eqref{conf:power-certificate}. Under
$P$, a minimum one-certificate is all one with probability $e^{-u}$,
which proves the final assertion.
\end{proof}

If $Q(f=1)\ge c$ for $0<c<1$, $H=\ln(1/c)$, $KS>H$, and
$t=2KS/u\ge1$, then
\begin{equation}\label{conf:certificate-cost}
 u\le\frac{6KS}{\ln(KS/H)}.
\end{equation}
Indeed, \eqref{conf:power-certificate} yields
$KS\,e^{-t}/t^2\le H$, so
$\ln(KS/H)\le t+2\ln t\le3t$.
When all $w_i\ge\ln2$, the condition $t\ge1$ is automatic, because
$u\le\sum_iw_i\le S/\ln2$.

\subsection{Gaussian testing with limited working memory}

\begin{lemma}[Testing bound in terms of the number of states]\label{conf:gaussian-test}
Let $u_1-u_0=h>0$, and let a randomized $K$-state test of $n$ independent
$N(u_i,1)$ observations have power at least $c\in(0,1]$ under $i=1$.
Define
\[
 A_h=2(\ln2)^2+2h^2(1+h^2/4),\qquad
 N(c)=\left\lceil\max\left\{e^2,
        \left(\frac{\ln(4/c)}{(\ln2)^2}\right)^2\right\}\right\rceil.
\]
For every $n\ge N(c)$,
\begin{equation}\label{conf:gaussian-bound}
 P_0(D=1)\ge\frac{c^2}{16}
    \exp\left\{-\frac{96A_h}{c}\frac{Kn}{\ln n}\right\}.
\end{equation}
The bound remains valid for independent Gaussian observations with
positive mean differences bounded by $h$.
\end{lemma}

\begin{proof}
Write $\xi=P_0(D=1)$. Equivalence of the experiments rules out $\xi=0$,
and $\xi=1$ is immediate. By Lemma~\ref{conf:monotone}, replace the test
by a deterministic monotone test with power at least $c/2$ and false
positive probability at most $2\xi/c$.

Center each observation at $m=(u_0+u_1)/2$ and write
$X_i-m=(2B_i-1)R_i$, where $R_i=|X_i-m|$. The vector of magnitudes has
the same law $\mathsf M$ under both hypotheses. Conditional on this
vector, the signs are independent, with
\[
 P_0(B_i=1\mid R_i=r_i)=\frac1{1+e^{hr_i}},\qquad
 P_1(B_i=0\mid R_i=r_i)=\frac1{1+e^{hr_i}}.
\]
For a fixed magnitude vector, the conditional test is a monotone binary
read-once program of width at most $K$. Conditioning only fixes the
transition functions for the analysis; it supplies no additional stored
information to the algorithm.

Set $w_i=\ln(1+e^{hr_i})$, $S(\mathbf r)=\sum_iw_i^2$, and let
$q(\mathbf r)$ be the conditional power of the monotone test. Since
$\mathbb E_{\mathsf M}q\ge c/2$ and $0\le q\le1$,
$\mathsf M\{q\ge c/4\}\ge c/4$. Also
\[
 w_i\ge\ln2,\qquad
 \mathbb E_{\mathsf M}w_i^2
 \le2(\ln2)^2+2h^2\mathbb E R_i^2=A_h.
\]
Markov's inequality therefore gives a set $\mathcal B$ of magnitude
vectors with mass at least $c/8$ on which
\[
 q(\mathbf r)\ge c/4,\qquad
 n(\ln2)^2\le S(\mathbf r)\le8A_hn/c.
\]
For a nonconstant conditional test on this set, apply
\eqref{conf:certificate-cost} with $H=\ln(4/c)$. The definition of
$N(c)$ implies $\ln(KS/H)\ge\tfrac12\ln n$. A minimum one-certificate
then has weight at most
\[
 u\le\frac{12KS}{\ln n}
       \le\frac{96A_h}{c}\frac{Kn}{\ln n}.
\]
Under hypothesis zero, this certificate occurs with probability
$e^{-u}$. The test that always returns one satisfies the same lower bound, and a
test that always returns zero cannot have the required conditional power.
Integrate over $\mathcal B$ using the common magnitude law, and then
use the upper bound $2\xi/c$ for the monotone test's false positive
probability. This gives \eqref{conf:gaussian-bound}.

For nonidentically distributed observations, use their respective
midpoints and differences $h_i\in(0,h]$. Monotone reduction still
applies, the magnitude law is common under the hypotheses, and the same
argument uses $\mathbb E w_i^2\le A_h$.
\end{proof}

\begin{lemma}[Effective sample size]\label{conf:effective}
For every fixed $c\in(0,1]$ there is $C_c<\infty$ such that, for
$0<h=u_1-u_0\le1$ and every $n,K\ge1$, a $K$-state Gaussian test with
power at least $c$ satisfies
\begin{equation}\label{conf:effective-bound}
 -\ln P_0(D=1)
 \le C_cK\frac{nh^2+1}{\ln(nh^2+1+e)}.
\end{equation}
\end{lemma}

\begin{proof}
Partition the observations into $N=\lceil n/m\rceil$ nonempty blocks of
size $m_j\le m=\lceil h^{-2}\rceil$, and let
$Z_j=m_j^{-1/2}\sum_{i\text{ in block }j}X_i$. These are independent
Gaussians with unit variance with mean differences $h\sqrt{m_j}\le\sqrt2$.
The original test can be simulated by a randomized $K$-state program
reading the $Z_j$'s. On receiving $Z_j$, generate independent standard
Gaussians $G_1,\ldots,G_{m_j}$ and temporarily reconstruct
\[
 X_i'=Z_j/\sqrt{m_j}+G_i-\overline G.
\]
Under either hypothesis, this vector has the required mean and identity
covariance; joint Gaussianity gives the original independent block
distribution. Execute its original transitions in temporary computation
and retain only the state at the end of the block. Thus the simulation preserves both
testing errors without increasing persistent memory.

When $N\ge N(c)$, Lemma~\ref{conf:gaussian-test}, with $h=\sqrt2$, gives
$-\ln P_0(D=1)\le C_cKN/\ln(N+e)$ after adjusting the constant. When
$N<N(c)$, binary data processing for relative entropy gives
\[
 \frac{nh^2}{2}
 \ge c\ln\frac1{P_0(D=1)}-\ln2.
\]
Here binary data processing follows by applying Jensen's inequality to
the likelihood ratio separately on the acceptance event and its
complement; the binary entropy is at most $\ln2$. Since $nh^2\le2N$
and $N$ is bounded in this case, the same bound follows with a larger
$C_c$. Finally, $N\le nh^2+1$ and $x/\ln(x+e)$ is increasing for
$x\ge1$, proving \eqref{conf:effective-bound}.
\end{proof}

\subsection{The joint risk lower bound and collapse}

\begin{theorem}[A joint confidence lower bound]
\label{thm:main-confidence-lower}
There is an absolute $c>0$ such that, for $0<\delta\le1/16$,
\[
\Rstr_{n,K,\delta}(1)
\ge c\min\left\{1,K^{-2}+\frac Ln+
\frac{L\log(e+L)}{nK}\right\}.
\]
The bound allows randomized procedures with transitions that vary over time and randomized final decoding.
\end{theorem}

\begin{proof}[Proof of Theorem~\ref{thm:main-confidence-lower}]
Fix a uniformly safe program and write
$r=\sup_{|\mu|\le1}\mathbb E_\mu(\widehat w-\mu)^2/2$.
The case $r\ge1/8$ is absorbed into the numerical constant. Otherwise
take $h=\sqrt{8r}<1$. The elementary statistical and quantization lower
bounds give $r\ge1/(32n)>0$ and $r\ge1/(16K^2)$.
For completeness, the first follows by testing $\mu=\pm1/(2\sqrt n)$:
their total variation is at most $1/2$, and an incorrect sign incurs
loss at least $1/(8n)$. For the second, replace each randomized terminal
output by its conditional mean given its state, which can only reduce
squared risk. The resulting codebook has at most $K$ points. Integrating
squared error to the nearest codepoint over the uniform distribution on
$[-1,1]$ gives risk at least $1/(6K^2)$, and hence the stated weaker
bound. This reduction is used only for the lower bound, so it need not
preserve safety.

Let $E=\{\widehat w\ne0\}$ and $s=\ln(1/\delta)$. Safety at $\mu=0$
implies $P_0(E)\le e^{-s}$. At $\mu=h$, a zero output incurs loss
$h^2/2=4r$, so $P_h(E)\ge3/4$. Changing the terminal decoder to
$\mathbf1_E$ yields a test with the same state budget. For the full
observations and all independent program randomization, the likelihood
ratio of mean $h$ against mean zero is
\[
 \Lambda=\exp\left\{h\sum_{i=1}^nX_i-\frac{nh^2}{2}\right\},
 \qquad \mathbb E_0\Lambda^2=e^{nh^2}.
\]
Cauchy--Schwarz gives
\[
 \frac34\le P_h(E)
 \le\{\mathbb E_0\Lambda^2P_0(E)\}^{1/2},\qquad
 nh^2\ge s-2\ln(4/3).
\]
Since $s\ge\ln16$, this implies $r\ge L/(32n)$ and
$x:=nh^2+1=8nr+1\ge s$. Lemma~\ref{conf:effective}, with $c=3/4$,
now yields
\[
 s\le CK\frac{x}{\ln(x+e)},\qquad
 x\ge\frac{s\ln(s+e)}{CK}=:z.
\]
The target term $T=L\ln(e+L)/(nK)$ is at most $4Cz/n$.
If $z\ge2$, then $r\ge z/(16n)\ge T/(64C)$. If $z<2$, use
$r\ge1/(32n)\ge T/(256C)$. Combining this with the quantization and
$L/n$ bounds proves
\[
 r\ge c_*\min\left\{1,K^{-2}+\frac Ln+
                   \frac{L\ln(e+L)}{nK}\right\}
\]
with a numerical $c_*>0$. The argument is uniform in the program, so
taking its infimum proves the theorem.
\end{proof}

The following form makes the consequence of insufficient state growth
explicit.
\begin{proposition}[Collapse to the baseline]\label{conf:collapse}
If $K_n\ge1$ and $\delta_n\in(0,1)$ satisfy
\begin{equation}\label{conf:collapse-condition}
 \frac{\ln(1/\delta_n)\ln n}{K_n n}\longrightarrow\infty,
\end{equation}
then, uniformly over all $K_n$-state uniformly $\delta_n$-safe programs,
$P_\mu(\widehat w\ne0)\to0$ for every fixed $\mu\in(0,1]$.
Consequently, $\Rstr_{n,K_n,\delta_n}(1)\to1/2$.
\end{proposition}

\begin{proof}
For any fixed power threshold $c>0$, apply
Lemma~\ref{conf:gaussian-test} to $D=\mathbf1\{\widehat w\ne0\}$ and
the mean pair $(0,\mu)$. Its constants are independent of the selected
program and of $K_n$. If its power were at least $c$, safety would imply
\[
 \ln(1/\delta_n)\le C_{\mu,c}\frac{K_nn}{\ln n}
                         +\ln(16/c^2),
\]
contradicting \eqref{conf:collapse-condition} for sufficiently large
$n$. Since $c$ is arbitrary, the asserted uniform convergence follows.
At $\mu=1$, risk is at least $P_1(\widehat w=0)/2$, whereas the
program that always returns zero has worst-case risk exactly $1/2$.
\end{proof}

\subsection{A concrete validation construction at fixed target risk}

At a fixed target risk, a pilot estimate of fixed resolution supplies a
finite collection of candidates. The construction retains the pilot
while reusing a workspace to validate its sign and magnitude.

\begin{lemma}[A local vote with adjustable accuracy]\label{conf:local-vote}
Fix $0<a\le1$, $w\in[a,\min(2a,1)]$, $\sigma\in\{-1,1\}$, and
$r\ge1$. There is a deterministic vote using at most $131r/a$ states
and $\lceil8192r/a^2\rceil$ observations such that
\[
 \begin{split}
 \sigma\mu\le w/2&\quad\Longrightarrow\quad P_\mu(Z=1)\le e^{-r},\\
 \sigma\mu\ge3w/4&\quad\Longrightarrow\quad P_\mu(Z=0)\le e^{-r}.
 \end{split}
\]
The parameters $w,\sigma$ are fixed inputs to this subroutine; storing
their labels will be charged separately.
\end{lemma}

\begin{proof}
Form increments $Y_i=2\mathbf1\{\sigma X_i\ge5w/8\}-1$, and run their
random walk from zero with absorbing boundaries
$\pm b$, where $b=\lceil64r/a\rceil$. After the fixed length
$m=\lceil8192r/a^2\rceil$, accept only if the positive boundary was
reached. There are $2b+1\le131r/a$ states.

Write $d=\mathbb E Y_i$. Under the two alternatives in the lemma,
respectively, $d\le-a/16$ and $d\ge a/16$, since
$2\Phi(a/8)-1\ge a/16$. For $d>0$, let
$\theta=\ln((1+d)/(1-d))\ge2d$. The process
$\exp(-\theta\sum_{i\le k}Y_i)$ is a nonnegative martingale.
Stopping at the first visit to $-b$ or at time $m$ bounds the probability
of reaching the wrong boundary by $e^{-2db}\le e^{-8r}$.
Reflecting the walk proves the corresponding bound for $d<0$.
Under $d\ge a/16$, outside the event of reaching the negative boundary,
nonacceptance implies that the unabsorbed walk ends below $b$.
Nonacceptance is therefore contained in the union of these two events.
As $md-b\ge447r/a$ and $m\le8193r/a^2$, Hoeffding's
inequality bounds this probability by
\[
 \exp\left\{-\frac{(md-b)^2}{2m}\right\}\le e^{-12r}.
\]
Thus the two errors are at most $e^{-8r}+e^{-12r}\le e^{-r}$.
Absorbed paths ignore the remaining observations until the prescribed
block boundary; disjoint blocks consequently yield independent votes.
\end{proof}

\begin{lemma}[Candidate validation at fixed resolution]\label{conf:validation}
Let $A=16384$ and let $B$ be the numerical constant for mean squared error in
Lemma~\ref{lem:scalar-engine}. Put $C_R=(25B+2)/2$.
For every fixed power of two $M=2^J\ge2$, every $0<\delta<1$ satisfying
$J\le\ln L$, and every $t\in[V,U]$, there is a uniformly
$\delta$-safe program with at most $2AMt$ states, at most
$C_NM^2LU/t$ observations, and worst-case risk at most $C_R/M^2$.
Here $C_N$ is numerical.
\end{lemma}

\begin{proof}
Use Lemma~\ref{lem:scalar-engine} to produce the unshrunk pilot
\[
 q=\Phi^{-1}\!\left(\Pi_{[\Phi(-1),\Phi(1)]}(k/M)\right),
 \qquad k\in\{0,\ldots,M-1\}.
\]
It uses at most $2AM$ states and $32AM^2$ observations and satisfies
$\mathbb E_\mu(q-\mu)^2\le B/M^2$. Candidates with $|q|<1/M$ are
replaced by zero. For the others, preassign consecutive validation
intervals to the amplitude buckets
\[
 a_j=2^j/M,\qquad |q|\in[a_j,2a_j),\qquad
 \beta_j=4^{-j}/16,\qquad 0\le j<J,
\]
including the endpoint $1$ in the last bucket. A path operates only in
its candidate's bucket and ignores the other validation intervals.

For a candidate in bucket $j$, put $w=|q|$, $\sigma=\operatorname{sgn}(q)$.
Apply Lemma~\ref{conf:local-vote} with accuracy parameter $r=4t$:
each fresh vote has errors at most $e^{-4t}$ between
$H_0:\sigma\mu\le w/2$ and $H_1:\sigma\mu\ge3w/4$.
Its state and sample costs are bounded by $524t/a_j$ and
$32769t/a_j^2$. With $s=\ln(1/\delta)$, set
\begin{equation}\label{conf:validation-schedule}
 R=\max\{1,\lceil s/(2t)\rceil\},\qquad
 k_j=\left\lceil\frac{\ln(R/\beta_j)}{4t}\right\rceil.
\end{equation}
Take the OR of $k_j$ fresh votes, repeat this on $R$ disjoint groups,
and accept the candidate only if all $R$ OR decisions are one.
The candidate is otherwise replaced by zero.

Because $t\le U\le L$, we have $R\le L/t$. The condition
$j<J\le\ln L$ gives
\[
 \ln(R/\beta_j)\le\ln L+\ln16+2j\ln2\le3U,
 \qquad 1\le k_j\le3U/t\le e^{2t},
\]
where the final inequality follows from $t\ge V=1+\ln U$.
Under $H_1$, an inner OR fails with probability at most
$e^{-4tk_j}\le\beta_j/R$, so the outer AND fails with probability
at most $\beta_j$. Under $H_0$, an inner OR succeeds with probability
at most $k_je^{-4t}\le e^{-2t}$, and independence gives total
acceptance probability at most $e^{-2tR}\le\delta$.

These bounds imply safety after conditioning on the pilot: outputting
$q\ne0$ is harmful precisely when $\sigma\mu<w/2$, and its conditional
acceptance probability is then at most $\delta$. Each candidate is
validated only once, so averaging over the pilot requires no union
bound over buckets.

For risk, let $e=q-\mu$. If $|q|<1/M$, the squared error after rejection
is at most $2e^2+2/M^2$. If $\nu=\sigma\mu<3w/4$, then
$|e|=|w-\nu|\ge w/4$ and $|\mu|\le5|e|$, so either accepting or
rejecting has squared error at most $25e^2$. Otherwise rejection has
probability at most $\beta_j$, and the conditional squared error is
at most
\[
 e^2+\beta_j\mu^2\le3e^2+8a_j^2\beta_j
                  =3e^2+\frac1{2M^2}.
\]
Averaging proves worst-case risk at most $(25B+2)/(2M^2)$.

We now charge all persistent information. A bucket contains at most
$4Ma_j$ pilot labels. For $a_j<1/2$, this follows because the two
Gaussian probability intervals corresponding to the positive and
negative buckets each have length at most $a_j$, and each contains at
most $Ma_j+1$ grid points. For the last bucket, the total number of
labels is simply at most $M\le4Ma_j$, including labels clipped to the
endpoints. Each active label is paired with the walk state and two
Boolean flags for the inner OR and outer AND. The total state count
during validation is therefore at most
\[
 (4Ma_j)\,4\,\frac{524t}{a_j}+M+1
     \le8386Mt<2AMt.
\]
The added $M+1$ states cover inactive labels and zero. Block indices,
bucket indices, and positions in the logical schedule are known from
the public clock. At a boundary, the same transition releases the
workspace and retains the candidate label. Thus pilot and validation
stages reuse the same alphabet; their state counts are not added.

Finally, the number of observations for bucket $j$ is at most
\[
 Rk_j\frac{32769t}{a_j^2}
 \le98307M^2\frac{LU}{t}\,4^{-j}.
\]
Summing the geometric series and adding $32AM^2$ pilot observations
gives the claimed numerical $C_N$. All block lengths and all
boundaries in \eqref{conf:validation-schedule} are fixed before
observing the data.
\end{proof}

\subsection{Matching tradeoffs at fixed risk}

\begin{proof}[Proof of Theorem~\ref{thm:main-confidence}]
For the lower bound, fix $\rho\in(0,1/2)$ and put $c=1-2\rho>0$.
Every uniformly safe program with worst-case risk at most $\rho$
gives the test $D=\mathbf1\{\widehat w\ne0\}$ satisfying
$P_1(D=1)\ge c$ and $P_0(D=1)\le e^{-s}$, where $s=\ln(1/\delta)$.
Cauchy--Schwarz with the Gaussian likelihood ratio gives
$n\ge s+2\ln c$. For $s$ sufficiently large depending only on $\rho$,
this implies $n\ge s/2\ge N(c)$. Lemma~\ref{conf:gaussian-test}
then gives
\[
 s\le C_{1,c}\frac{Kn}{\ln n}+\ln(16/c^2),\qquad
 n\ge\frac{s\ln n}{2C_{1,c}K}.
\]
Increasing the same lower threshold on $s$ ensures
$s\ge L/2$ and $\ln n\ge U/2$. Hence
\[
 n\ge L/4,\qquad n\ge LU/(8C_{1,c}K),
\]
which proves the lower bound for every $K$.

For the upper bound, choose a power of two
\[
 M_\rho\ge\max\{2,\sqrt{2C_R/\rho}\},\qquad
 D_\rho=2AM_\rho.
\]
Decrease $\delta_\rho$ if needed so that
$\log_2M_\rho\le\ln L$ whenever $\delta\le\delta_\rho$.
For $K\ge D_\rho V$, let
\[
 t=\min\{U,K/D_\rho\}\in[V,U].
\]
Lemma~\ref{conf:validation} supplies a safe program with at most
$D_\rho t\le K$ states, risk at most $\rho/2$, and observation count
at most
\[
 C_NM_\rho^2\frac{LU}{t}
  =C_NM_\rho^2L\max\{1,D_\rho U/K\}
  \le C_\rho L\max\{1,U/K\}.
\]
Absorb integer rounding into $C_\rho$ to obtain the upper bound.
\end{proof}

\begin{proof}[Proof of Corollary~\ref{cor:main-exponential}]
The preceding test reduction at means zero and one has fixed power
$c=1-2\rho$. For $\delta_n=e^{-\gamma n}$,
Lemma~\ref{conf:gaussian-test} forces
\[
 \gamma n\le C_{1,c}\frac{Kn}{\ln n}+\ln(16/c^2).
\]
Thus $K\ge c_\rho\gamma\ln n$ for all sufficiently large $n$.

For sufficiency, apply Theorem~\ref{thm:main-confidence} with
$L_n=1+\gamma n$ and $U_n=1+\ln L_n$. Choose $K_n$ to be the smallest
power of two at least $\max\{U_n,D_\rho V_n\}$. Then
$K_n=O_{\rho,\gamma}(\ln n)$, and the theorem's construction uses at
most $C_\rho L_n$ observations. Choose
$\gamma_0(\rho)\le1/(2C_\rho)$. For every fixed
$0<\gamma\le\gamma_0(\rho)$ and sufficiently large $n$, its required
sample count is at most $n$. Ignore any remaining observations.
Together with the lower bound, this gives
$K_{\min}=\Theta_{\rho,\gamma}(\ln n)$ and hence
$b_{\min}=\log_2\ln n+O_{\rho,\gamma}(1)$.
\end{proof}

\subsection{Confidence costs in fixed dimension}
\label{app:fixed-d-confidence}

\begin{proof}[Proof of Corollary~\ref{cor:main-fixed-d-confidence}]
\emph{Reduction for the lower bound.}
Restrict a $d$-dimensional procedure to means $\mu=\theta e_1$,
$|\theta|\le1$. Given a scalar observation $Y_t\sim N(\theta,1)$,
generate an independent $Z_t\sim N_{d-1}(0,I_{d-1})$ using fresh
randomness and feed $(Y_t,Z_t)$ to the procedure. For $d=1$, this is
the original scalar observation. The simulated procedure keeps the same
$K$ persistent states. If its vector output is $W$, return $W_1$.
Pointwise,
\[
\tfrac12(W_1-\theta)^2\le\tfrac12\|W-\theta e_1\|^2,
\qquad R_{\theta e_1}(0)=\theta^2/2.
\]
The scalar unsafe event is therefore contained in the vector unsafe
event, and scalar risk is bounded by vector risk. Thus
$N_\rho^{(d)}(K,\delta)\ge N_\rho(K,\delta)$, and
Theorem~\ref{thm:main-confidence} supplies the lower bound for every $K$.

\emph{Sequential validation and its state count.}
Let $A,C_R,C_N$ be the constants in Lemma~\ref{conf:validation}.
Choose a fixed power of two
\[
M=2^J\ge\max\{2,\sqrt{2dC_R/\rho}\},\qquad
q=M+1,\qquad S=2AMq^{d-1}.
\]
The construction in that lemma outputs either zero or one of the $M$
pilot candidates
\[
q_k=\Phi^{-1}\!\left(\Pi_{[\Phi(-1),\Phi(1)]}(k/M)\right),
\qquad 0\le k<M.
\]
Hence at most $q$ output labels suffice, independently of the confidence
level. Put
\[
L_d=1+\ln(d/\delta),\qquad U_d=1+\ln L_d,\qquad V_d=1+\ln U_d.
\]
For any $t\in[V_d,U_d]$ with $J\le\ln L_d$, run the lemma's
construction at confidence $\delta/d$ on coordinate $j$, using a
predetermined block of
\[
m=\left\lceil C_NM^2L_dU_d/t\right\rceil
\]
fresh vector observations. It reads coordinate $j$ from each observation
and ignores the others. Pad each coordinate's schedule to $m$ observations
by retaining its final label.

During block $j$, the completed coordinates occupy at most $q^{j-1}$
labels, and the active scalar procedure has at most $2AMt$ states.
Their product is at most $St$. At the block boundary, the scalar output
is encoded as its candidate label and the next scalar procedure is
initialized in the released workspace. The final vector has at most
$q^d\le St$ labels, since $q\le2AM$ and $t\ge1$. The public clock
identifies the coordinate block and the local schedule. Thus all stages
can be embedded in one alphabet of at most $St$ states; no block index
or separate copy of the workspace is retained. The total sample count is
\begin{equation}\label{eq:fixed-d-resources}
N=dm\le d(C_N+1)M^2L_dU_d/t.
\end{equation}
Here $M^2L_dU_d/t\ge1$ absorbs the rounding of $m$.

Each scalar parameter lies in $[-1,1]$. The resulting vector therefore
satisfies
\[
\sup_{\mu\in B_2^d}\E_\mu R_\mu(W)
\le\sum_{j=1}^d C_R/M^2\le\rho/2.
\]
If every coordinate's loss is at most its loss under the zero baseline, the sum
has the same property. A union bound consequently gives
\[
\sup_{\mu\in B_2^d}\Pp_\mu\{R_\mu(W)>R_\mu(0)\}
\le\sum_{j=1}^d\delta/d=\delta.
\]

\emph{Matching the prescribed budget.}
Choose $\delta_{d,\rho}\le\delta_\rho$ small enough that
$L\ge\max\{1,\ln d,e^J\}$. Then $L_d=L+\ln d\le2L$,
$U_d\le2U$, $V_d\le2V$, and $J\le\ln L_d$.
Set $D_{d,\rho}=2S$. Whenever $K\ge D_{d,\rho}V$, choose
\[
t=\min\{U_d,K/S\}\in[V_d,U_d].
\]
The state count is at most $St\le K$, including the final output labels.
Equation~\eqref{eq:fixed-d-resources} gives
\begin{align*}
N&\le d(C_N+1)M^2L_d\max\{1,SU_d/K\}\\
 &\le4d(C_N+1)M^2S\,L\max\{1,U/K\}.
\end{align*}
This proves the upper bound in \eqref{eq:fixed-d-confidence} with constants
depending only on $d,\rho$. The actual integer states embed directly in
the prescribed alphabet $K=2^b$.

\emph{Exponential confidence.}
Write $H=d(C_N+1)M^2$ and choose $\gamma_0(d,\rho)=1/(2H)$.
For fixed $0<\gamma\le\gamma_0(d,\rho)$ and $\delta_n=e^{-\gamma n}$,
take $t=U_d$ and let $K_n$ be the smallest power of two at least $SU_d$.
For all sufficiently large $n$, the lemma's conditions hold and
\[
N\le H(1+\ln d+\gamma n)\le n,
\qquad K_n<2SU_d=O_{d,\rho,\gamma}(\ln n).
\]
After the $N$ scheduled observations, keep the final label and ignore the
remaining observations. Conversely, the lower bound already proved gives,
for every admissible $K$ and all sufficiently large $n$,
\[
n\ge c_\rho(1+\gamma n)\frac{1+\ln(1+\gamma n)}K,
\qquad
K\ge c_\rho\gamma\,[1+\ln(1+\gamma n)].
\]
Together these inequalities yield the asserted state and bit asymptotics.
\end{proof}

\subsection{Terminal compression at high confidence}

\begin{proposition}[Terminal compression at high confidence]
\label{prop:conf-terminal}
For every $n\ge1$, $K=2^b$, and $0<\delta<1$,
\[
 \mathcal R^{\mathrm{term}}_{n,K,\delta}(1)
 \le8\min\{1,L/n+K^{-2}\}.
\]
Consequently, for each fixed $\rho\in(0,1/2)$, a fixed alphabet size
$K_\rho$ suffices for terminal risk at most $\rho$ under
$\delta_n=e^{-\gamma n}$, for all sufficiently large $n$ and every fixed
$0<\gamma\le\rho/32$.
\end{proposition}

\begin{proof}
Quantize the projection of $\overline X$ onto $[-1,1]$ to the nearest
midpoint of $K$ equal intervals, obtaining $q$ with
$|q-\mu|\le|\overline X-\mu|+K^{-1}$ and
$\mathbb E_\mu(q-\mu)^2\le2/n+2K^{-2}$.
Decode its label by soft thresholding at
$\lambda=\sqrt{2\ln(2/\delta)/n}+K^{-1}$.
The Gaussian tail bound makes $|q-\mu|\le\lambda$ hold with probability
at least $1-\delta$, so Lemma~\ref{lem:radial-safety}'s deterministic
safety argument applies. Its risk calculation gives
\[
 \mathbb E_\mu R_\mu(\widehat w)
 \le\frac{2+4\ln(2/\delta)}n+4K^{-2}
 \le8(L/n+K^{-2}).
\]
Choosing the zero action when its bound is smaller proves the minimum.
Finally choose a power of two $K_\rho\ge\sqrt{32/\rho}$.
For $\gamma\le\rho/32$, the displayed bound is at most
$8/n+\rho/2$, which is at most $\rho$ for sufficiently large $n$.
\end{proof}

\end{document}